\documentclass[lettersize,journal]{IEEEtran}
\usepackage{graphics} 
\usepackage{graphicx}
\usepackage{epsfig} 
\usepackage{url}
\usepackage{xcolor}
\usepackage{amsmath}
\usepackage{amssymb}
\usepackage{multirow}

\newcommand{\argmin}{\mathop{\mathrm{argmin}}}

\newcommand{\degree}{$^\circ$}

\newcommand{\fnew}{f_{new}} 
\newcommand{\pnew}{p_{new}} 
\newcommand{\tnew}{t_{new}} 
\newcommand{\pest}{p_{est}} 
\newcommand{\fanc}{f_{a}} 
\newcommand{\panc}{p_{a}} 
\newcommand{\fancA}{f_{a1}} 
\newcommand{\fancB}{f_{a2}} 
\newcommand{\pancA}{p_{a1}} 
\newcommand{\pancB}{p_{a2}} 

\IEEEoverridecommandlockouts                              

\title{\LARGE \bf
CoPR: Towards Accurate Visual Localization With Continuous Place-descriptor Regression}

\author{Mubariz Zaffar$^{1}$, Liangliang Nan$^{2}$ and Julian F. P. Kooij$^{3}$
\thanks{*This work was supported by the 3D Urban Understanding (3DUU) lab funded by the TU Delft AI Initiative.}
\thanks{$^{1,2}$ Authors are with the Intelligent Vehicles Group,  Department of Cognitive Robotics, 
Delft University of Technology (TU Delft), 2628CD Delft, The Netherlands
        {\tt\small \{m.zaffar,j.f.p.kooij\}.tudelft@nl}}%
\thanks{$^{3}$Author is with the 3D Geoinformation Group, Faculty of Architecture and the Built Environment,
Delft University of Technology (TU Delft),
2628BL Delft, The Netherlands
        {\tt\small liangliang.nan@tudelft.nl}}%
}

\begin{document}
\maketitle
\thispagestyle{empty}
\pagestyle{empty}

\begin{abstract}
Visual Place Recognition (VPR) is an image-based localization method that estimates the camera location of a query image by retrieving the most similar reference image from a map of geo-tagged reference images. In this work, we look into two fundamental bottlenecks for its localization accuracy: reference map sparseness and viewpoint invariance.
Firstly, the reference images for VPR are only available at sparse poses in a map, which enforces an upper bound on the maximum achievable localization accuracy through VPR.
We therefore propose Continuous Place-descriptor Regression (CoPR) to densify the map and improve localization accuracy.
We study various interpolation and extrapolation models to regress additional VPR feature descriptors from only the existing references.
Secondly, we compare different feature encoders and show that CoPR presents value for all of them. We evaluate our models on three existing public datasets and report on average around 30\% improvement in VPR-based localization accuracy using CoPR, on top of the 15\% increase by using a viewpoint-variant loss for the feature encoder.
The complementary relation between CoPR and Relative Pose Estimation is also discussed.
\end{abstract}


\section{Introduction}
\IEEEPARstart{O}{ne} of the key research problems for robotics and computer vision is accurate Visual Localization (VL), i.e., 
to localize a robot in a map using as input only an image from the robot's camera~\cite{toft2020long}.
Various parallel research directions have emerged within VL.
A top-level distinction can be made between purely image-based approaches and 3D structure-based approaches.
The former are simple and efficient but have lower localization accuracy, while the latter are more accurate at the cost of increased computation complexity and maintenance effort~\cite{piasco2018survey}.
Purely image-based approaches could be further divided into Visual Place Recognition (VPR)~\cite{arandjelovic2016netvlad}, Absolute Pose Regression (APR)~\cite{kendall2015posenet}, and Relative Pose Estimation (RPE)~\cite{laskar2017camera}.
Given their efficiency and scalability, VPR techniques are often used in robotics for loop closure detection or 3D reconstruction. However improving their performance remains an ongoing research challenge~\cite{lowry2015visual} \cite{zaffar2021vpr}.

In VPR the task is to find for a query image the best matching reference image from
a set of pre-recorded geo-tagged reference images (i.e., the reference map) 
~\cite{garg2021your}.
Each reference image is considered a `place',
and the geo-location of the best-matched reference
is then the estimated location (`place') of the query image.
Whereas VPR relies on image-retrieval,
in APR a neural network directly regresses the global coordinates
for a query image,
and the map is implicitly represented by the network weights. 
However, such APR methods do not generalize across viewpoints, as has been studied by Sattler et al.~\cite{sattler2019understanding}.
RPE on the other hand 
operates on two images with assumed nearby viewpoints, and estimates from the overlapping image contents the relative translation and orientation between their corresponding camera coordinate frames.
Since VPR performs coarse global localization, and RPE performs fine-grained localization by assuming coarse localization is solved, both techniques are often combined in the multi-stage approach, referred to as Coarse-to-Fine localization (CtF)~ \cite{laskar2017camera} \cite{balntas2018relocnet} \cite{ding2019camnet}. 
RPE is therefore not an alternative to VPR, but a refinement step that is only successful if VPR was able to retrieve a nearby reference.

\begin{figure*}[htbp]
\begin{center}
\includegraphics[width=1.0\linewidth]{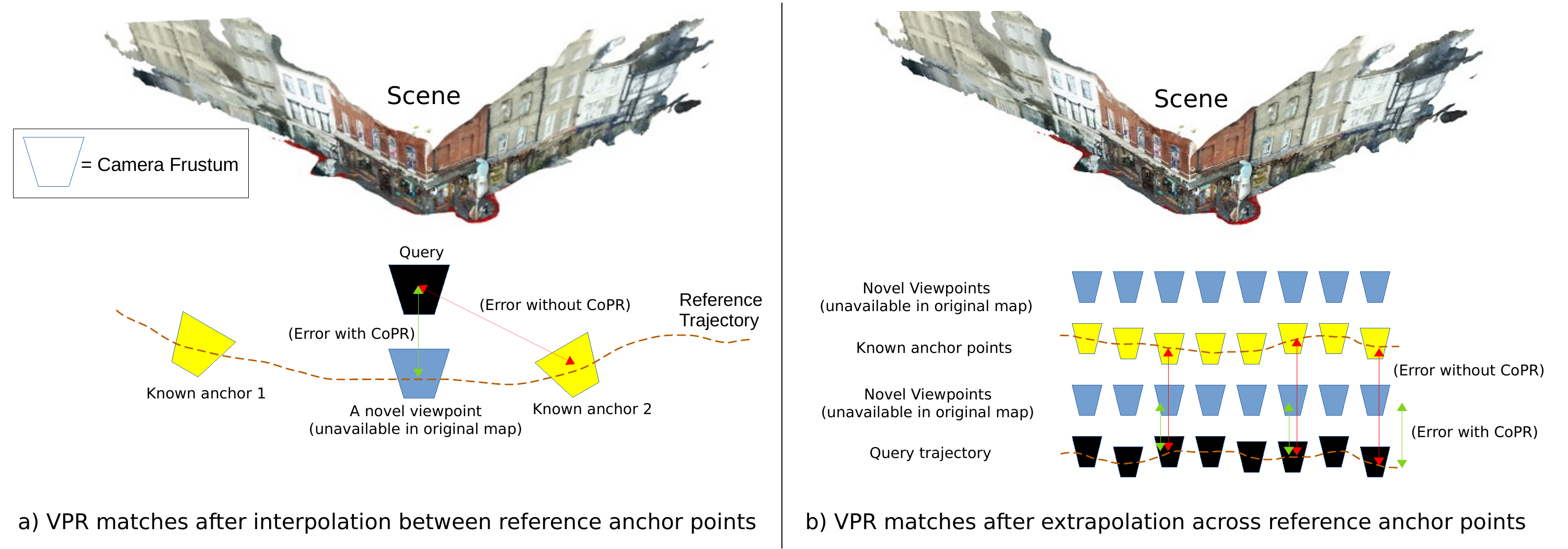}
\end{center}
\caption{The discrete treatment of VPR that leads to lower localization accuracy. Provided that only the \textit{yellow} anchor reference poses are available in the map, the \textit{black} query images could only be matched as close as possible to the base error. Regressing descriptors for the \textit{blue} target viewpoints using interpolation or extrapolation given anchor reference descriptors could lead to improved localization accuracy for query images in VPR and thus reduce the base error. The scene shown in this figure is taken from the work of Sattler et al.~\cite{sattler2019understanding}.}
\label{Fig:CoPR_intuitive_explanation}
\end{figure*} 

VPR remains less accurate than structure-based and CtF approaches~\cite{sattler2018benchmarking}, with
a crucial reason being the discrete nature of the reference map in VPR.
When a query image appears between two anchor locations in the reference map, a VPR system could at best only match this to the nearest spatial anchor location, incurring some minimal Euclidean distance error. This can become worse when query images and existing reference images span the same area but at offsets of parallel lines, as shown in Fig.~\ref{Fig:CoPR_intuitive_explanation}.
Therefore, we seek to add more references to the map (such as the blue poses in Fig.~\ref{Fig:CoPR_intuitive_explanation}),
a notion referred to as \textit{map densification}.
A trivial but often impractical solution to densification is by collecting more reference images.
Alternatively, densification could be achieved by creating a 3D model of the environment and rendering images at novel poses. However, creating and maintaining up-to-date 3D models is computationally and storage-wise expensive, and the resulting images are not photo-realistic~\cite{sattler2019understanding, moreau2022lens}.

Since the VPR reference maps comprise compact feature descriptors of images, we suggest performing map densification in the \textit{feature space} rather than the image space.
We propose Continuous Place-descriptor Regression (CoPR) in feature space for VPR map densification\footnote{This discrete nature of reference map is also problematic for APR as reported by~\cite{sattler2019understanding}. We hypothesize that APR could also benefit from map densification via descriptor regression, but this aspect is not explored in this work and we limit its scope to VPR.}.
Since in CtF the RPE step assumes the initial VPR  step was performed correctly, we note that
improving VPR could also address CtF errors that cannot be corrected by RPE, as we will also show in this work.

We argue for two requirements to benefit from such map densification: 1) a method of regressing meaningful feature descriptors for VPR at novel target viewpoints given anchor point feature descriptors, 2) an image-retrieval system that is viewpoint-variant and therefore could utilize the regressed descriptors at target viewpoints.
Furthermore, the model for descriptor regression should only need existing anchor descriptors and relative poses between anchor locations and target viewpoints, at its input, and it should not require images of the scene from target viewpoints or expensive scene reconstruction~\cite{sattler2019understanding}.

To study the problem of descriptor regression, we further consider two possible schemes: interpolation and extrapolation. Both of these are relevant for map densification, where \textit{interpolation} (Fig.~\ref{Fig:CoPR_intuitive_explanation}a) refers to interpolating to an intermediate location between some anchor points on the reference trajectory, while \textit{extrapolation} (Fig.~\ref{Fig:CoPR_intuitive_explanation}b) refers to regressing descriptors around a given anchor reference pose. Since interpolation could even be performed using averaging of the nearest anchor points along the trajectory, i.e., by simply following the trend in the local feature space, we expect it to be an easier problem to solve than extrapolation. 
Extrapolation, on the other hand, is a more important requirement for map densification, because it enables us to potentially regress descriptors at or close to the query. Interpolation can at best only densify within an existing reference trajectory. 

Finally, for a VPR system to benefit from map densification, it needs to retrieve the Euclidean closest match in the physical space as the best match in the feature space. This is not enforced in  VPR techniques trained with triplet-loss~\cite{arandjelovic2016netvlad}, classification-loss~\cite{chen2017deep} and ranking-based-loss~\cite{revaud2019learning}, where the correct/incorrect ground-truth match is discrete (leading to viewpoint invariance), instead of the continuous ground-truth in distance-based loss~\cite{thoma2020geometrically}.
If a VPR technique is viewpoint-invariant, both the blue trajectories in Fig. \ref{Fig:CoPR_intuitive_explanation}b would be incorrectly considered equally valid. Thus, we hypothesize that map densification and viewpoint variance should work hand in hand to make VPR-based localization more accurate.
We
show that a highly viewpoint-variant VPR technique in a densified reference map leads to the highest localization accuracy, amongst all the combinations originating from the different feature encoders and levels of map densification. \\

In summary,
our contributions are as follows:

\begin{enumerate}

    \item We investigate Continuous Place-descriptor Regression (CoPR) to densify a sparse VPR map through either interpolation or extrapolation of the feature descriptors to target poses, without requiring any new measurements (i.e., reference images).

    \item We propose linear regression based techniques and a non-linear deep neural network for map densification and demonstrate the improvement in localization accuracy on
    three existing public datasets.

    \item We report that different feature encoders can benefit from map densification and the best performance is achieved by using the most viewpoint-variant descriptors in a densified map.
        
    \item We discuss the VPR failure cases where RPE cannot recover the correct pose without CoPR, highlighting the complementarity of these approaches for improving VL accuracy. We demonstrate the existence of such cases with real-world data.
\end{enumerate}

\section{Related Work}
In this section, we expand on the existing body of literature for Visual Localization, as reviewed in \cite{piasco2018survey}: a system that consists of retrieving the pose (position + orientation) of a visual query material within a known space representation. Such systems are further classified into direct and indirect methods. The direct methods consist of Absolute Pose Regression (APR), structure-based localization, and Coarse-to-Fine localization (CtF). The indirect methods are: Visual Place Recognition (VPR) approaches, which is a robotics problem, and image-retrieval, which is a computer vision problem. Both of these mostly represent the same formulation but with a few differences regarding evaluation metrics and experimental setup as discussed in~\cite{zaffar2021vpr}. In this research, our scope is limited to VPR-based localization and its limitations, however, to understand these limitations and due to the significant overlap between various fields of VL and the collective benefit from map densification, we expand on all these fields in the following.

\textbf{Structure-based approaches:} 
These approaches use 2D-3D matching given 2D pixels and 3D scene coordinates to yield highly accurate pose estimates. Recent benchmarks~\cite{sattler2018benchmarking, torii2019large, sattler2019understanding} have shown that such structure-based approaches are state-of-the-art when it comes to accurate localization. The work of Li et al.~\cite{li2012worldwide} is seminal in this field that shows large-scale structure-based localization by proposing a co-occurrence prior to RANSAC and bidirectional matching of image features with 3D points. Efficiency is of importance and Liu et al.~\cite{liu2017efficient} propose the use of global contextual information derived from the co-visibility of 3D points for 2D-3D matching. InLoc~\cite{taira2018inloc} presents a formulation for structure-based localization in indoor environments by using dense feature matching for texture-less indoor scenes and view synthesis for verification. Active-Search~\cite{sattler2016efficient} uses 2D-to-3D and 3D-to-2D matching for pose regression and candidate filtering, while DSAC++~\cite{brachmann2018learning} uses learnt scene-coordinate regression building upon DSAC~\cite{brachmann2017dsac}. Both of these techniques form the state-of-the-art for structure-based 6-DoF camera localization~\cite{sattler2019understanding}. While structure-based approaches are highly accurate, they require significant computations and have limited scalability, and maintaining and updating the corresponding potentially large-scale 3D models is challenging. 

\textbf{Absolute Pose Regression:} 
APR started from the seminal works of PoseNet~\cite{kendall2015posenet} and the incremental build-up by authors in \cite{kendall2016modelling} and \cite{kendall2017geometric}, and has since seen many different variants of it e.g., the works in ~\cite{naseer2017deep, valada2018deep, clark2017vidloc, walch2017image}. The objective of APR approaches is to memorize an environment given a set of images and their corresponding ground-truth poses, such that given a new image, the network can generalize from the poses seen at training time and directly regress the new pose. In \cite{melekhov2017image}, an encoder-decoder architecture is employed with a final regressor network to regress camera pose. Radwan et al.~\cite{radwan2018vlocnet++} present a multi-task learning framework for visual-semantic APR and odometry. While APR methods are simple and efficient, they have been shown to suffer from degeneralization across viewpoints and appearances, and are unable to extrapolate to parallel trajectories~\cite{sattler2019understanding}. 

\textbf{Coarse-to-Fine localization:}
Another approach to the problem of accurate localization is a two-staged Coarse-to-Fine formulation, where the first stage is VPR and the second stage is Relative Pose Estimation (RPE). This need for CtF approaches arises because the query trajectories and reference trajectories are usually far apart, and the coarse VPR stage can only at best retrieve the closest pose on the reference trajectory. Thus, there is always a base error in the coarse VPR stage, which is then reduced by the RPE module for fine-grained localization. Authors in~\cite{laskar2017camera} propose a CtF approach by using a Siamese network architecture for RPE. RelocNet~\cite{balntas2018relocnet} uses camera frustum overlap information at training time, while CamNet~\cite{ding2019camnet} models the CtF localization approach in three separate modules with increasing fineness. The work of \cite{saha2018improved} models pose estimation by discovering and computing relative poses between pre-defined anchor locations in the map. Most of these CtF approaches model RPE as a pose regression problem given global descriptors leading to a lack of scene generalization. Thus authors in PixLoc~\cite{sarlin2021back} instead learn local features useful for geometric 2D-3D matching which can generalize to new scenes. SANet~\cite{yang2019sanet} also models the CtF localization pipeline using 2D-to-3D matching by learning scene coordinate regression and generalizes to new scenes. However, both of these approaches require a coarse 3D model of the environment at their inputs.
 
\textbf{VPR and image-retrieval:}
VPR and image-retrieval in essence represent the same problem: i.e., given a query image and a map of reference images, retrieve the Nearest Neighbor (NN) reference matches for that query image. Depending on whether the closest match is required (VPR) or all of the possible matches need to be retrieved (image-retrieval), the problem favours loop-closure or 3D modelling, as discussed in \cite{zaffar2021vpr}. In this work, we use the two terminologies interchangeably to refer to the same problem. Both these tasks are usually treated as viewpoint-invariant and trained with losses such as triplet-loss~\cite{arandjelovic2016netvlad, gordo2017end, radenovic2018fine}, classification-loss~\cite{chen2017deep} and ranking-based-loss~\cite{revaud2019learning}. These losses aim to align the feature representation for viewpoint-varied images of the same place, which explicitly favours viewpoint-invariance. On the other hand, more recent distance-based loss functions explicitly force the network to encode geometric information within the feature descriptors, such that the top-most retrieved images are also the geometrically-closest images~\cite{thoma2020geometrically} \cite{thoma2020soft}. For our work, such a distance-based loss is highly relevant, since map densification could offer more benefit to VPR-based localization using viewpoint-variant feature descriptors than viewpoint-invariant descriptors.

Before dedicated datasets were developed for VPR, off-the-shelf Convolutional Neural Network (CNN) features were utilised, Chen et al.\cite{chen2014convolutional} used features from the Overfeat Network \cite{sermanet2014overfeat} and combined them with the spatial filtering scheme of Seq-SLAM. The use of off-the-shelf features of AlexNet trained on ImageNet for VPR was studied by Sunderhauf et al.~\cite{sunderhauf2015performance}, who found that some layers were most robust to conditional variations than others. Chen et al.~\cite{chen2017deep} proposed two neural networks, namely AMOSNet and HybridNet, which were trained specifically for VPR on the Specific Places Dataset (SPED). 

Recently, contrastive learning has been the dominant trend in VPR, as shown in \cite{arandjelovic2016netvlad} \cite{radenovic2018fine} but which classifies a place as the same or different in a hard (0/1) manner i.e., an image is considered as either the same place or a different place. But with multiple viewpoint-varied images of the same place, such a hard distinction is not possible and a soft distinction is required. For this purpose, the authors in~\cite{leyva2021generalized} present the concept of generalised contrastive loss based on image-content overlap. Previously discussed distance-based loss functions can also be classified as soft losses since they can distinguish between multiple viewpoint-varied images of the same place. Other than this, VPR literature includes the use of ensembles of VPR techniques to reject false positives~\cite{hausler2019multi, hausler2020hierarchical}. 

\textbf{Implicit scene representations:}
In addition to the concept of explicit 3D models for structure-based approaches, implicit scene representation has been more popular recently, where the structure is stored within the parameters of a neural network. Such implicit scene representation could come from neural implicit representations~\cite{mescheder2019occupancy} \cite{oechsle2019texture}, differentiable volumetric rendering~\cite{niemeyer2020differentiable} or the more recent trends in Neural Radiance Fields (NeRF)~\cite{mildenhall2020nerf}. If the structure is known, whether implicitly or explicitly, it is possible to synthesize images at new viewpoints of the scene. These synthesized images could be directly used for map densification in a VPR-based localization system~\cite{torii201524}, for pose verification in a Coarse-to-Fine localization system~\cite{zhang2021reference} or for creating more training data for absolute pose regression approaches~\cite{moreau2022lens}. Authors in~\cite{yen2021inerf} invert the NeRF process to refine the camera pose estimate given an initial coarse estimate. However, implicit scene representation approaches offer similar challenges as structure-based approaches for localization regarding maintaining and updating the scene representations. They also suffer from scalability and artifacts created in the image space, as reported in \cite{moreau2022lens}. 

In summary, VPR is an efficient and easy-to-maintain localization method compared to structure-based approaches, it is more generalizable than APR techniques and simpler than multi-staged CtF approaches; however, it remains less accurate than CtF and structure-based approaches, where this accuracy is related to the sparseness of the reference map at creation and viewpoint-variance of the feature encoder. One possibility to increase this localization accuracy as surveyed here is to use the CtF approaches in a retrieval-followed-by-regression manner, however, this itself depends on the quality of the initial coarse retrieval stage, i.e., VPR, such that an incorrectly retrieved coarse estimate leads to a definite failure of the complete CtF pipeline. 

Therefore, we instead look in a different direction than CtF and explore some of the fundamental reasons for the inaccuracy of VPR. We investigate whether it is possible to increase VPR-based localization accuracy even without relying on RPE as a second stage and without requiring any additional measurements of the scene. For this, we look into densifying the map of descriptors and the benefits of such map densification for different types of VPR feature encoders. 

\section{Methodology}
In this section, we first provide an overview of our problem statement. We then dedicate sub-sections to introduce the concept of map densification (CoPR), the descriptor regression strategies for CoPR, and the different feature encoders for VPR. Lastly, we discuss the relationship between CoPR and RPE.

\subsection{Problem statement}
\label{methodologyoverview}
Given a set of reference images with known poses, VPR constructs a map $M = (R,P)$, where $R$ is a set of reference descriptors, such that $f_i \in R$ is an $N$-dimensional feature descriptor with a corresponding pose $p_i \in P$.
Each feature descriptor $f_i  = G(I_i)$ is obtained from a reference image $I_i$ using an already trained and fixed feature extractor $G$, typically a neural network.
The pose $p_i$ is a 6 degree-of-freedom pose that specifies the location as a translation vector $t_i = (x, y, z)$, and a quaternion vector $o_i$ specifying the 3D orientation.

At test time, the objective is to find the pose $p_q$ of a query image $I_q$, for which the query descriptor $f_q = G(I_q)$ is computed. 
The descriptor $f_q$ is matched to all the reference descriptors in the set $R$, and the Nearest Neighbor (NN) match $r_{nn} = \argmin_{r \in R}|| f_r - f_q ||_2$ is retrieved. The pose of the query image is then considered the same as that of the retrieved reference descriptor, i.e., $p_q=p_{nn}$.
Ideally, the feature descriptors are constructed such that the resulting Euclidean translation error $e=||t_q-t_{nn}||_2$ is minimal.
Hence, the assumption $p_q=p_{nn}$ is essentially an approximation $p_q \approx p_{nn}$, and would only be true in the unlikely event that the query  is collected at the same pose as that of the retrieved
reference in the map.
Thus, the expected error $\mathbb{E}[e]$ is a non-zero \textit{base error} of a VPR system.
This base error is directly affected by the sparseness in the reference map: the further apart the reference samples are, the higher the base error could be\footnote{Clearly, if the query images appear at the exact same spot as that of the reference trajectory, map densification would not help. This however is highly unlikely and unrealistic in real-world situations as evident in existing VPR datasets.~\cite{zaffar2021vpr}}. Therefore, this work proposes to apply map densification for VPR as shown in Fig.~\ref{Fig:CoPR_intuitive_explanation}. 

\subsection{Map densification}
To reduce the base error, we seek to extend the number of descriptors and poses in a given sparse map $M_{sparse}$. 
Since collecting more reference images is not always possible,
we aim to perform densification 
using only existing reference descriptors in $M_{sparse}$ without the need to collect more images at novel viewpoints. Such densification in feature space also has computational benefits since image-description is more computationally expensive than descriptor-regression, as shown later in sub-section~\ref{computational_details}. 
Concretely, we propose to densify a sparse map $M_{sparse}=(R,P)$ by defining a set of target poses $P'$ for which the corresponding descriptors $R'$ are predicted via Continuous Place Descriptor Regression (CoPR) using one or more existing reference descriptors in $R$ which we will refer to as \textit{anchor descriptors}.
The resulting densified map $M_{dense}=(R \cup R', P \cup P')$ thus extends the original map $M_{sparse}$ with the newly regressed target references. 

Different strategies could be employed to define (a) which set of target poses $P'$ to regress to, and (b) how to regress the descriptors for a target pose using the available anchor descriptors.
We here explore two specific strategies for defining the set $P'$, namely
(1)
interpolating between the anchor points on the reference trajectory.
and (2)
extrapolating to nearby poses of an anchor pose that do not necessarily lie along the reference trajectory.
Regression approaches will be discussed later in  sub-section~\ref{sec:descriptorregressor}.

The \textbf{interpolation scheme}
assumes that the references in the sparse map are obtained in a sequence. 
Additional poses $P'$ can be selected along the trajectory in between the poses available in $P$.
Hence, any two subsequent references $a1 \in R$ and $a2 \in R$ can be selected as anchors, and one or more new target poses $\pnew$ can be selected on the path between the anchor poses $\pancA$ and $\pancB$.

In the \textbf{extrapolation scheme}
the set of target extrapolation poses $P'$ are selected in the vicinity of the poses in $P$, but not necessarily on a path between them.
One possibility is to generate these target poses in a uniform grid within a certain distance threshold around each anchor.
Another possibility is to define a single global uniform grid, and only evaluate grid points using the nearest anchor points (within some distance threshold) similar to \cite{sattler2019understanding}.
The former approach leads to a denser grid, although it is globally non-uniform. 

\begin{figure}[htb]
\begin{center}
\includegraphics[width=1.0\linewidth]{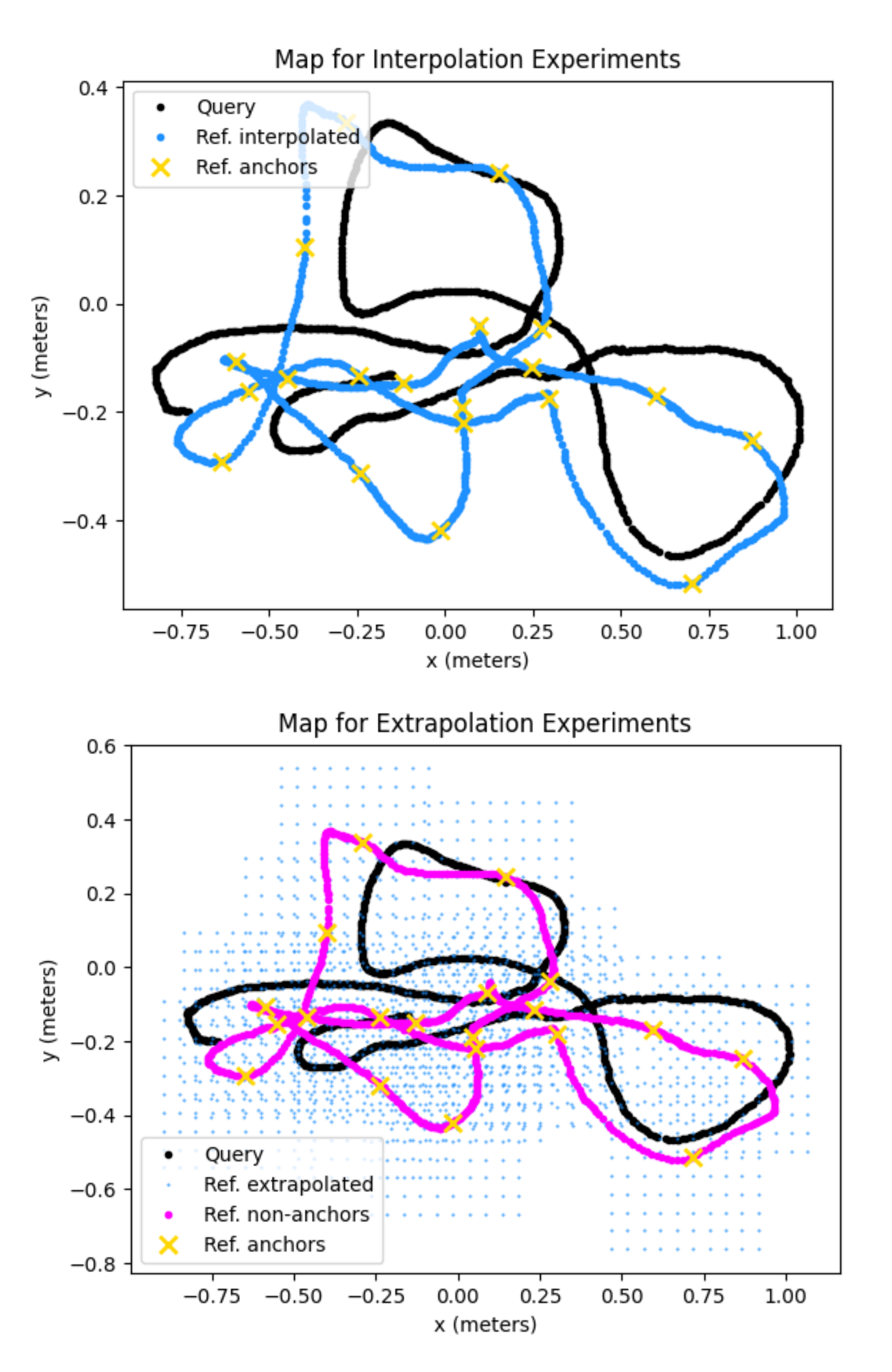}
\end{center}
\caption{The test setup for the interpolation and extrapolation experiments on the Heads scene of the 7-scenes dataset in 2D. The anchor reference points are to be used by regression techniques to interpolate/extrapolate descriptors at target poses. Since in the case of extrapolation we do not sub-sample along the reference trajectory as in interpolation, there are non-anchor reference points in the extrapolation experiment but not in the interpolation experiment.}
\label{Fig:interpolationandextrapolation_examples7scenes}
\end{figure} 

Examples of the reference, query, and target poses are shown  in Fig.~\ref{Fig:interpolationandextrapolation_examples7scenes} to illustrate interpolation and extrapolation for map densification on the 7-scenes dataset~\cite{glocker2013real}.

\subsection{Descriptor regression strategies}
\label{sec:descriptorregressor}
We consider several strategies
to predict a new descriptor $\fnew{} \in R'$ for a given target pose $\pnew{} \in P'$ and the sparse reference map $M_{sparse}$,
which could be applied to the extrapolation and/or interpolation tasks. In principle, a regression method fits a model to express the dependent variable(s) as a function of the independent variables, thereby capturing the local trend in the space around the fitted samples. For feature descriptor regression, our objective is to express the feature space as a function of the pose. Since this feature space is latent, it is unclear to what extent we can assume it to be globally or locally linear for changing pose, hence we consider both linear and non-linear regression techniques for CoPR, as follows.

\subsubsection{\textbf{Linear interpolation}}
\label{sec:weightedlinearinterpolation}
The simplest strategy only applies to interpolation, where we only use the translation and not the orientation of each pose. 
We aim to predict the descriptor for an intermediate translation between two known translations.
The target descriptor in this case
is a linear weighted combination of its two anchors,
\begin{align}
    \fnew &=(1-\alpha_{a1}) \times \fancA + (1-\alpha_{a2}) \times \fancB, \\
    \alpha_{a1} &= \beta_1  \; / \; (\beta_1 + \beta_2), \\
    \alpha_{a2} &= \beta_2  \; / \; (\beta_1 + \beta_2),
\end{align}
where $\beta_1=||t_{new}-t_{a1}||_2$,  $\beta_2=||t_{new}-t_{a2}||_2$, and  $\fancA$, $\fancB$ are the two anchor feature descriptors.

\subsubsection{\textbf{Linear Regression using local plane fit}}
\label{sec:linearregressionusinglocalplanefit}
\begin{figure}[htbp]
\begin{center}
\includegraphics[width=1.0\linewidth]{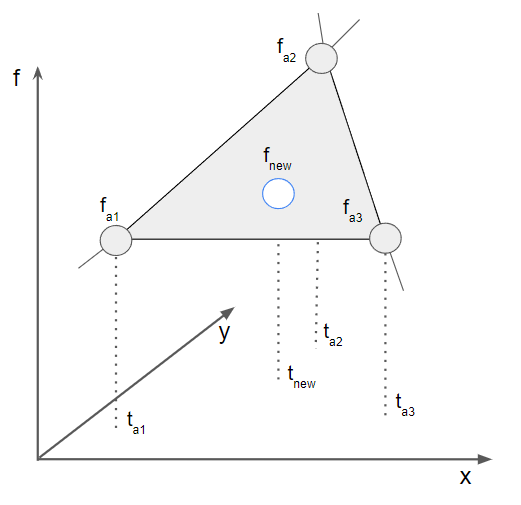}
\end{center}
\caption{A locally-fit plane given three anchor points in a two-dimensional world. Note that this plane is for a single feature dimension, so in practice there will be $N$ such planes.}
\label{Fig:linearregression_fig}
\end{figure}

As a second approach, 
we investigate a local plane fit
to consider more anchors and allow extrapolation too. This also only uses the translation and not the complete pose.
Given the target translation $\tnew{}$, the O Nearest Neighbor anchor points from $M_{sparse}$ in terms of Euclidean translation distance are selected.
For each descriptor dimension, a linear plane is least-squares fitted on the anchor values,
and the plane is evaluated at the translations of the target $\tnew$ to regress $\fnew$. This linear regression is abstractly depicted in Fig.~\ref{Fig:linearregression_fig} for a single feature dimension ($f$) in a two-dimensional pose space ($x$ and $y$).
Note that a more complex polynomial or spline regression 
could be used too, 
but we limit our approach to linear regression here as the most canonical implementation of this general approach.

\newcommand{\relpose}{\Delta p}
\subsubsection{\textbf{Non-linear regression network}}
\label{sec:nonlin-reg-network}
\begin{figure}[htbp]
\begin{center}
\includegraphics[width=1.0\linewidth]{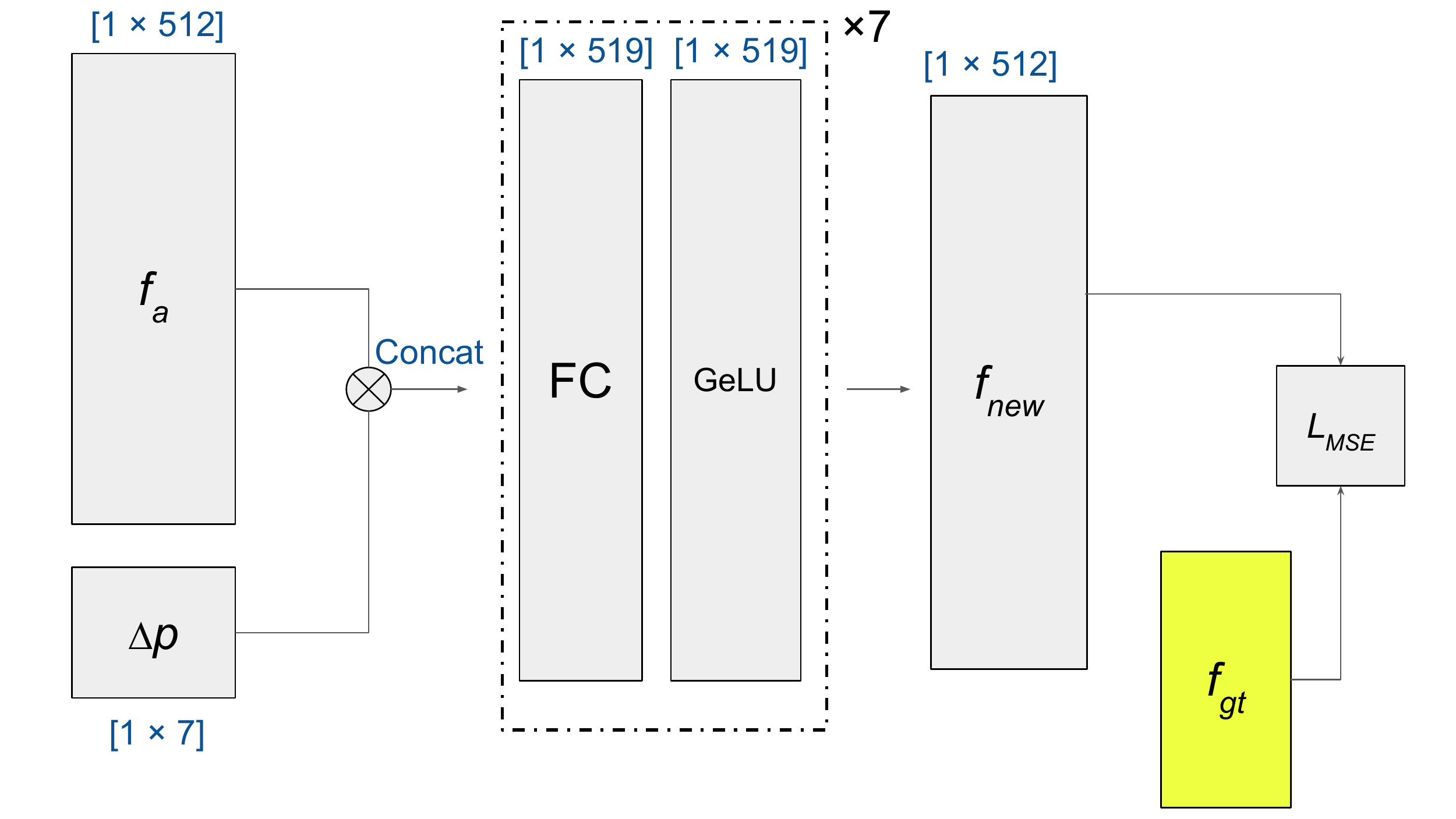}
\end{center}
\caption{The non-linear deep learning based model $H$ that we train to regress the descriptor $\fnew$ at a target location. The input is an anchor reference descriptor $\fanc$ and the relative pose $\relpose$ between the anchor location $\panc$ and the target location $\pnew$. }
\label{Fig:CoPR_model}
\end{figure}

In this strategy, we directly regress $\fnew = H(\fanc, \relpose)$ from a single anchor descriptor $\fanc$, and the relative pose $\relpose$ specifying the translation difference and the quaternion rotation between the anchor pose $\panc$ and the target pose $\pnew$.
As non-linear descriptor regressor $H$, we use a fully-connected deep neural network consisting of 7 hidden layers with a GeLU~\cite{hendrycks2016gaussian} activation. The input to the network is the $N$ dimensional anchor feature descriptor $\fanc$ and the relative pose $\relpose$ stacked together,
while the output is the $N$ dimensional target feature descriptor $\fnew$ at the pose $\pnew$. The dimensionality of the input layer and hidden layers is the same, i.e., $N+7$, as the relative pose vector $\relpose$ has a length of $7$, while the output layer has only $N$ dimensions.
This network is shown in Fig.~\ref{Fig:CoPR_model}.
In preliminary experiments on Microsoft 7-scenes (see sub-section~\ref{sec:experimentalsetup}) we explored other activations and using fewer or more layers. We found GeLU works best, and that the network can overfit with more than 7 layers.

Given a pre-trained and fixed encoder $G$ for computing feature descriptors, the non-linear regression network is trained on available descriptor pairs (e.g., an anchor descriptor $\fanc$ and a ground-truth target descriptor $f_{gt}$) with known relative pose $\relpose$ between them, and a mean-squared error loss,

\begin{equation}
L_{MSE}=||H(\fanc, \relpose) - f_{gt}||_2.
\end{equation}

\subsection{Losses for the feature encoder}
\label{sec:featureencoder}
Next, we discuss the choice for the training loss of the feature encoder $G$, since 
the feature space is key for the general localization quality, and also defines the complexity of the regression task that map densification should solve.
The feature encoder $G$ takes as input an image $I$ and computes its $N$ dimensional feature descriptor $f_I$.
We will compare three different training strategies, namely training with a triplet loss~\cite{arandjelovic2016netvlad}, an RPE loss~\cite{laskar2017camera}, and a distance-based loss~\cite{thoma2020geometrically}, which are 
shortly summarized here.

For training with a \textbf{triplet loss}, the network computes $N$ dimensional feature descriptors $\{f_q, f_p, f_n\}$ for three images $\{I_q, I_p,I_n\}$: a query $I_q$, a positive match $I_p$ with varied viewpoint and a negative match $I_n$ that represent a different scene/place. Each of these three $N$ dimensional feature descriptors is then normalized and penalized with a triplet loss. The triplet loss is the same as that of \cite{arandjelovic2016netvlad} which penalizes the network given a Euclidean distance function $d_f(f_1,f_2)=||f_1-f_2||_2$ and a margin $m$ with a triplet loss,
\begin{equation}
L_{triplet} = max\{d_f(f_q, f_p)-d_f(f_q, f_n)+m,0\}.
\end{equation}

For the \textbf{RPE loss}~\cite{laskar2017camera}, $f_q$ and $f_p$ are stacked together and passed through a relative-pose regressor consisting of fully-connected layers to output the estimated 6-DoF relative pose $\Delta \pest{}$ between the two input images. The network is trained with a mean-squared error loss, i.e.
\begin{equation}
L_{relative}=||\Delta \pest{}-\Delta p_{gt}||_2, 
\end{equation}

\noindent given the ground-truth relative pose $\Delta p_{gt}$.  This is the same network as that of Laskar et al.~\cite{laskar2017camera}. To regress the relative pose $\Delta \pest{}$ correctly, the network has to encode viewpoint information in the feature descriptors $\{f_q, f_p\}$. Nevertheless, this relative pose-based loss does not explicitly force the network to encode representations that 
encourage the closest descriptor in 3D physical space to be the closest in feature space.

Therefore, the third loss is the \textbf{distance-based loss} $L_{distance}$ as introduced in the work of Thoma et al.~\cite{thoma2020geometrically},
\begin{equation}
L_{distance} = ||\Delta f-\Delta t||_2.
\end{equation}
This loss explicitly penalizes the network based on the Euclidean distance $\Delta f$ between feature descriptors $\{f_q, f_p\}$ and the Euclidean distance between their corresponding ground-truth translation poses $\Delta t$.

\subsection{Relating CoPR to Relative Pose Estimation}
\label{sec:relatingcoprwithctf}
Our main focus is the task of VPR for VL. Nevertheless, map densification can also improve the accuracy of Coarse-to-Fine localization, i.e., ~VPR plus RPE~\cite{laskar2017camera}.
This sub-section expands on the methodological relation between CoPR and RPE.

Formally, given two feature descriptors $f_1$ and $f_2$ and the relative pose between their corresponding locations $\Delta p$, a CoPR strategy as in sub-section~\ref{sec:nonlin-reg-network}
models a function $f_2=H(f_1,\Delta p)$. In contrast, RPE aims to learn a function $\Delta p=L(f_1,f_2)$. While these two functions $H$ and $L$ appear similar, these approaches have different benefits. A useful property of CoPR is that it can be done offline, thus localization reduces to a single-stage image-retrieval problem at runtime, while RPE is performed online and thus leads to a multi-stage CtF formulation.

\begin{figure}[t]
\begin{center}
\includegraphics[width=1.0\linewidth]{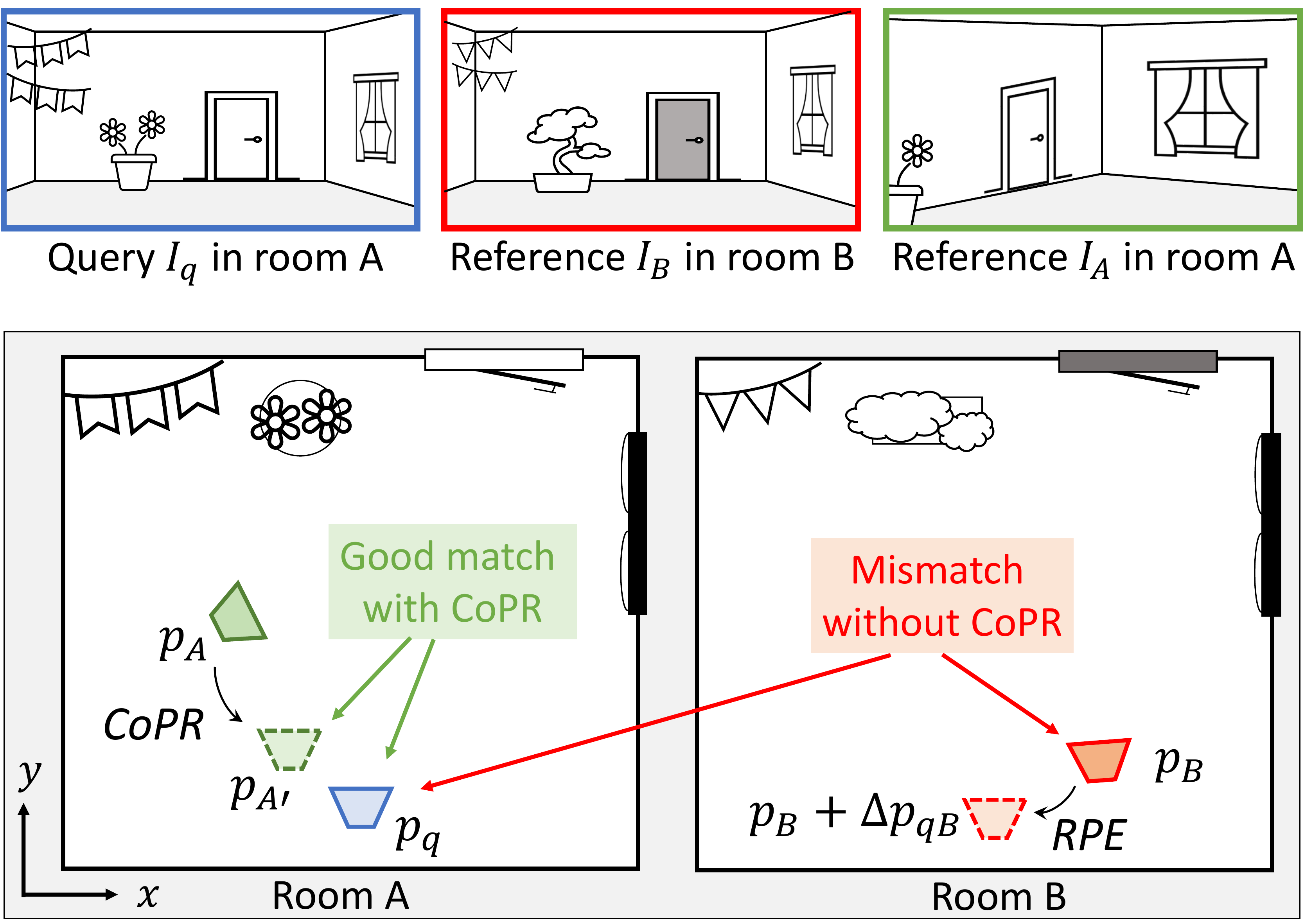}
\end{center}
\caption{Perceptual aliasing of rooms A and B: query $I_q$ in room $A$ appears more similar to reference $I_B$ in room $B$ than to reference $I_A$ in correct room $A$.
If VPR retrieves the wrong reference $f_B$ for $f_q$, RPE between $f_B$ and $f_q$ cannot correct this: the `apparent' difference between the query pose $p_q$ and reference pose $p_B$ is nearly zero.
CoPR therefore aims to improve VPR instead by adding references for more diverse poses to the map, e.g.~$f_{A'}$ for $p_{A'}$.
}
\label{Fig:perceptualaliasingexample}
\end{figure} 

A more crucial difference is that RPE assumes its two input images represent the same scene,
and thus must rely on the accuracy of the preceding image-retrieval step.
Consider a query $I_q$ taken in a scene $A$, e.g., a room in an office, and a sparse reference map containing various visually similar scenes, e.g., other rooms in the same office (see Fig.~\ref{Fig:perceptualaliasingexample}).
The image-retrieval system might fail and retrieve a reference $f_B$ from an arbitrarily distant scene (`room') $B$ instead of any nearby reference $f_A$ from the actual scene $A$, i.e.~when $||f_B - f_q||_2 < ||f_A - f_q||_2$.
We refer to the inability to distinguish such similar scenes as \textit{perceptual aliasing}~\cite{brachmann2019expert}.
These scenes should ideally all be represented as nearby references in the feature space,
but in a sparse reference map some scenes could be underrepresented,
and retrieving the best (or even top-$k$) matches for a query might never include the correct scene.
RPE cannot correct such retrieval failures.
For instance, a pose difference between correct reference $I_A$ and query $I_q$ (both at room $A$) could limit the visual overlap between their images, making their descriptors $f_A$ and $f_q$ dissimilar.
If the visual content of $I_b$ and $I_q$ appear more similar,
their pose difference would \textit{appear} relatively small, even though these are at completely different scenes.
Since $\Delta p_{qB} = L(f_q,f_B)$ will just estimate the small \textit{apparent} pose offset, RPE results in an incorrect final pose estimate for the query, $p_B + \Delta p_{qB}$.

By densifying the reference map, we can instead extend the references in room $A$ to represent more diverse poses.
A regressed descriptor $f_{A'}$ at a new pose $p_{A'}$ closer to the query than the original reference $p_A$ can improve the best match, $||f_{A'} - f_q||_2 < ||f_B - f_q||_2$,
resulting in a good VPR localization estimate $p_q \approx p_{A'}$.
We demonstrate the existence of this effect using constructed failure cases in our experiments of sub-section~\ref{sec:DescriptorRegressionCanBenefitRelativePoseEstimation}.
In CtF localization, RPE afterward still reduces this gap further by estimating $\Delta p_{qA'} = L(f_q,f_{A'})$, such that $p_q = p_{A'} + \Delta p_{qA'}$.
CoPR and RPE are therefore complementary techniques.

\section{Experiments}
In this section, we present our experimental setup in detail, including the datasets, baselines, and evaluation metrics.
First, we validate using the encoder $G_{distance}$ as our primary encoder. We then present our results of using descriptor regression for interpolation and extrapolation experiments. We show how different feature encoders can benefit from CoPR and the effect of map density on localization performance. We also show the relation between CoPR and CtF localization, and finally provide the computational details of our work.

\subsection{Experimental setup}
\label{sec:experimentalsetup}
Here we explain the datasets, evaluation metrics and the various parametric choices used in our experiments.
\subsubsection{Datasets}
We use three datasets for evaluation, Microsoft 7-scenes, the Synthetic Shop Facade,
and the Station Escalator dataset. Our choice of these datasets is based on their wide adoption for evaluating VL in existing literature as reviewed previously and their complementary nature: indoor vs outdoor, different levels of spatial coverage and different types (parallel vs intersecting) of traversals. We discuss each dataset in turn.

\textbf{Microsoft 7-scenes} dataset~~\cite{glocker2013real} has been a long-standing public benchmark for 6-DoF indoor localization~\cite{laskar2017camera, sattler2019understanding, sarlin2019coarse}. This dataset consists of seven different indoor scenes collected using a Kinect RGB-D camera and provides accurate 6-DoF ground-truth poses computed using a KinectFusion~\cite{newcombe2011kinectfusion} baseline. Each scene spans an area of a few square meters and contains multiple sequences/traverses (viewpoint-varied) within a scene. Each sequence itself then contains between 500 to 1000 images, where each image has a $640 \times 480$ pixels resolution. There are separate query and reference sequences which contain novel viewpoints of the same scene. The images and poses in the query trajectory act as our training set for training both the feature encoder $G$ and the non-linear descriptor regressor $H$. The reference trajectory is further divided into two splits: validation and test sets, with 40\% images in the validation set and 60\% images in the test set. The validation set is used for validating the encoder $G$ and the non-linear regression network $H$ at training time. This reference trajectory is then used for the interpolation and extrapolation experiments.

The \textbf{Synthetic Shop Facade dataset} proposed by~\cite{sattler2019understanding} represents images and poses regressed from a 3D model of a real-world outdoor shopping street~\cite{kendall2015posenet} and consists of multiple sequences/traverses of a single scene. It contains about 9500 images at novel viewpoints with an image resolution of $455 \times 256$ pixels. There are separate splits for query and reference sequences that contain different viewpoints. The training, validation and test sets follow the same strategy as that of the 7-scenes dataset.

The \textbf{Station Escalator dataset} proposed by~\cite{sattler2019understanding} contains two parallel trajectories through a station and is hence useful for studying extrapolation benefits across parallel lanes. The dataset contains 330 query images and 330 reference images with an image resolution of $1557 \times 642$ pixels and 6-DoF accurate poses. For this dataset, we intend to regress descriptors from one trajectory (say A) to its parallel trajectory (say B), thus the non-linear regression network $H$ needs to be trained with such relative pose change between A and B. Therefore, given the two original parallel trajectories, we divide both into three parts: training, validation and test sets. The training images are selected as every 50th image in both trajectories, while the remaining images are equally divided between the validation and test sets. The training images from both traverses are used to train the descriptor regression models. For experiments, the validation and test images from trajectory A combined together act as our query images. The validation and test images from trajectory B in addition to the training images from trajectory A act as the reference images. 

\subsubsection{Evaluation metrics}
The evaluation metric is the Median Translation Error (MTE) in meters and the Median Rotation Error (MRE) in degrees over all the estimated query images' poses, as commonly used in existing literature~\cite{laskar2017camera} \cite{sattler2019understanding} \cite{sarlin2019coarse}. 
The median is normally preferred over the mean since outliers can skew the latter by any amount.
The translation error is the Euclidean distance between the query image's translation and the best-matched reference image's translation. The rotation error is the angular difference between the quaternion vectors of a query image and its best-matched reference image, as used in the reviewed literature.

\subsubsection{Training details and parametric choices}
We use the output of the final global average pooling layer of a ResNet34~\cite{he2016deep} backbone feature encoder, and thus a feature descriptor size of $N=512$ is used throughout this work. The feature encoder $G$ and the non-linear descriptor regressor $H$ are trained separately. For training all the three feature encoders $G_{triplet}$, $G_{relative}$ and $G_{distance}$ and for non-linear regression network $H$, we use the Adam optimizer for model optimization with learning rates of $1e^{-5}$, $1e^{-4}$, $5e^{-5}$ and $5e^{-4}$ for $G_{triplet}$, $G_{relative}$, $G_{distance}$ and $H$, respectively. The weights of the ResNet34 backbone are initialized via pretraining on ImageNet-1K and fine-tuned on the datasets used in this work, while the non-linear regression network $H$ is trained from scratch for each dataset.

For training the encoder $G_{triplet}$, images from the training sets of different scenes of the 7-scenes dataset are chosen randomly to act as negatives, while images from the same scene with varied viewpoints are chosen as positives. We use a margin of $m=0.3$ for the triplet loss, same as \cite{arandjelovic2016netvlad}. The feature encoder $G$ is trained jointly on the training pairs of all the seven scenes in the 7-scenes dataset. The encoders trained using triplet loss ($G_{triplet}$) and RPE loss ($G_{relative}$) are only trained on the 7-scenes dataset and used for experiments on all the datasets, while the model trained using distance-based loss ($G_{distance}$) is trained separately for each dataset. We later show the reasons behind this separate training for distance-based loss in sub-section \ref{sec:EncoderLossfunctionandLocalizationAccuracy}. 

A dedicated non-linear regression model $H$ is trained for each of the three datasets. The non-linear regression model $H$ trained for one dataset is used for both the interpolation and extrapolation experiments of that dataset. For the least-squares plane fit to linearly regress each feature dimension, O=4 is chosen as the number of NN anchors, which is the minimum number needed to fit a plane in 4D (i.e.,~3D world plus 1D feature).

\subsection{Encoder loss function and localization accuracy}
\label{sec:EncoderLossfunctionandLocalizationAccuracy}
\begin{figure}[htbp]
\begin{center}
\includegraphics[width=1.0\linewidth]{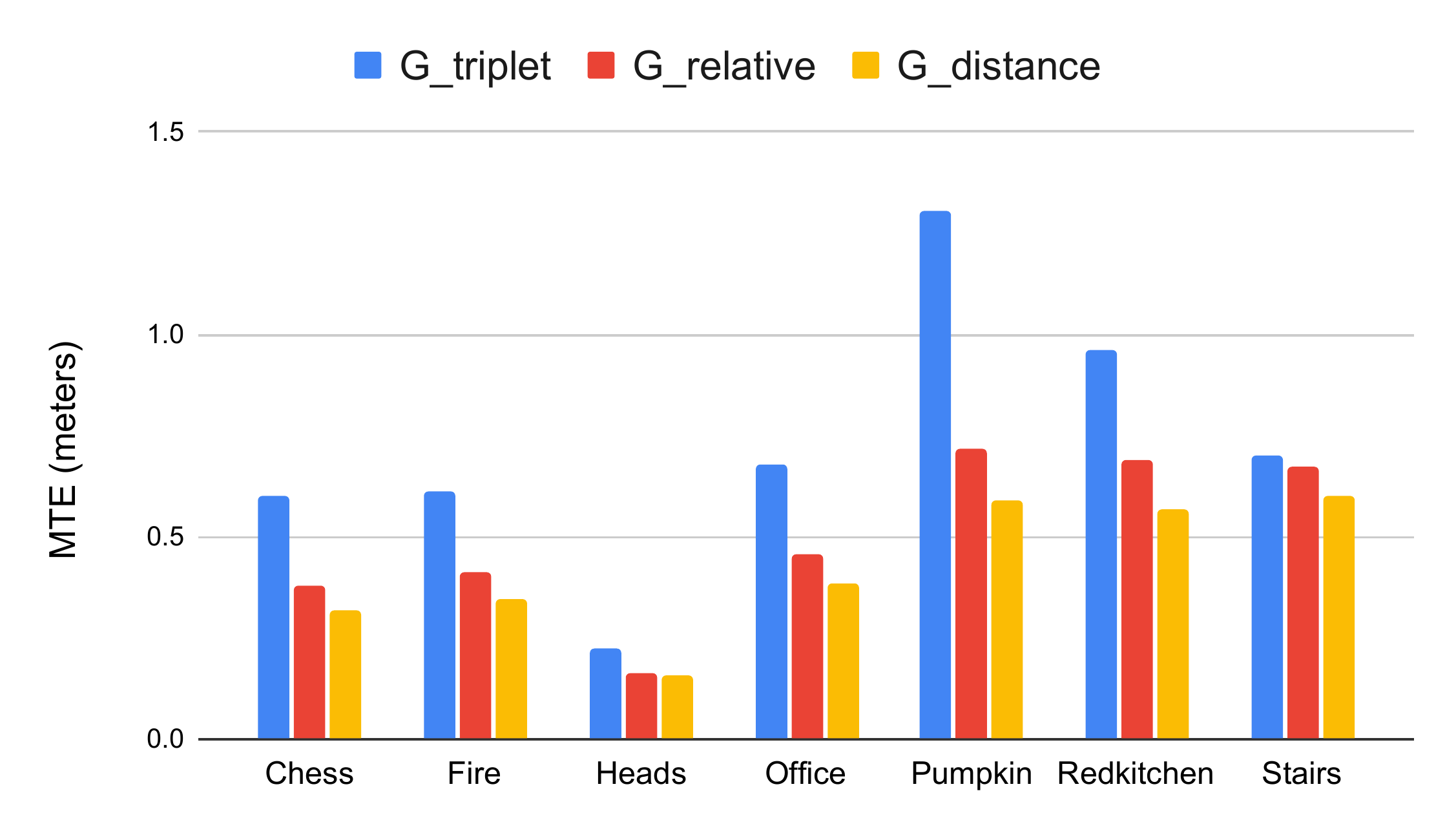}
\end{center}
\caption{The MTE of the three encoders when used for performing VPR-based localization on all the scenes of the 7-scenes dataset. Training with distance-based loss leads to lower MTE than other losses.}
\label{Fig:MedianTranslationErrorforDifferentFeatureEncoders}
\end{figure}

Here we intend to understand the first part of the two potential requirements for accurate VPR-based localization: viewpoint variance. The encoder training objectives favouring viewpoint variance can have a considerable effect on the VPR-based localization error. The change in localization error for $G_{triplet}$, $G_{relative}$ and $G_{distance}$ is shown in Fig. \ref{Fig:MedianTranslationErrorforDifferentFeatureEncoders} for the 7-scenes dataset, where a distance-based loss leads to the lowest localization error. This localization error is without map densification and is purely the effect of different training objectives for the encoder $G$. 

\begin{figure}[htbp]
\begin{center}
\includegraphics[width=1.0\linewidth]{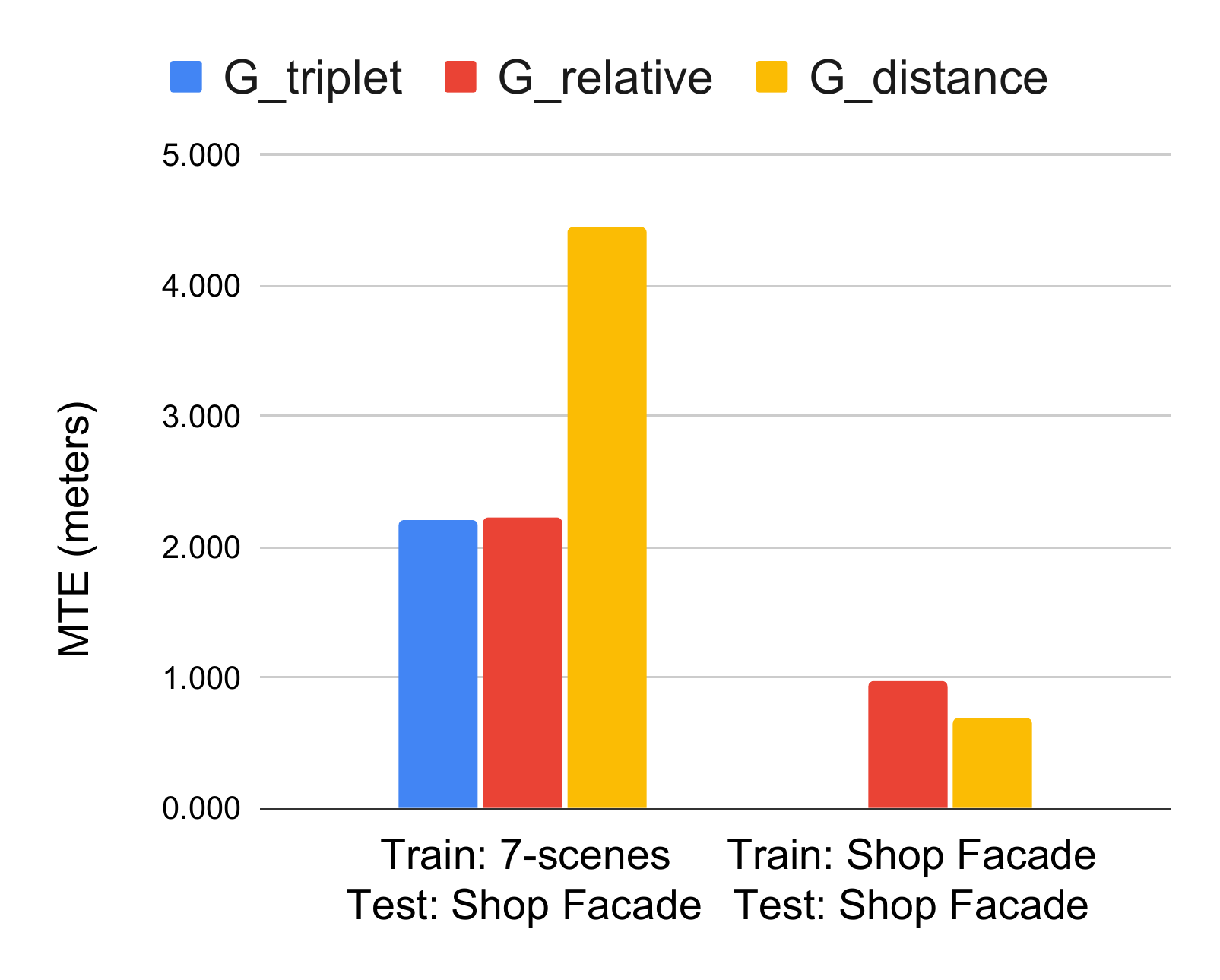}
\end{center}
\caption{The MTE of the three encoders used for testing VPR-based localization on the Synthetic Shop Facade dataset, when trained on the same and different dataset. Notably, $G_{triplet}$ and $G_{relative}$ trained on the 7-scenes can outperform $G_{distance}$ trained on the 7-scenes dataset. However,  $G_{distance}$ when trained and tested on the Synthetic Shop Facade dataset performs the best. Since the Shop Facade dataset contains images of only one scene, unlike the 7-scenes dataset, we could not select proper negative images in this dataset and do not train $G_{triplet}$ on this dataset.}
\label{Fig:DegeneralizationofDistancelossbasedEncoder}
\end{figure}

Moreover in Fig. \ref{Fig:DegeneralizationofDistancelossbasedEncoder}, we observe the (de)generalization of these feature encoders from one dataset to the other. This is done by evaluating the VPR-based localization performance of a given encoder on datasets other than the training dataset for a given model. We note that the network $G_{distance}$ trained on the 7-scenes dataset does not perform well on the Shop Facade dataset and is outperformed by $G_{triplet}$ and $G_{relative}$ trained on the 7-scenes dataset, which suggests that $G_{distance}$ is less generalizable. We therefore train $G_{distance}$ on the Shop Facade dataset, after which it outperforms the other networks. This degeneralization of distance-based loss has also been reported by \cite{thoma2020geometrically} and an intuitive explanation could be that distance-based losses are more sensitive to structural changes between different domains and the change in scene appearance with changing scene depth.

Since distance-based loss leads to the lowest localization error, we only use $G_{distance}$ as our backbone encoder for the experiments in sub-sections \ref{sec:extrapolationexperiments} and \ref{sec:interpolationexperiments}. However, we later show in sub-section \ref{sec:MapdensificationwithDifferentFeatureEncoders} that all the encoders ($G_{triplet}$, $G_{relative}$ and $G_{distance}$) can benefit from CoPR, albeit at varying levels of accuracy.

\subsection{Extrapolation experiments}
\label{sec:extrapolationexperiments}

We first explain the setup used for extrapolation experiments, followed by the extrapolation methods and baselines, and then the corresponding results and discussion.

\subsubsection{Extrapolation setup}
We use all three datasets to examine the effects of extrapolation. All of these three datasets have properties useful for our CoPR analysis. Thus, we first explain the setup for extrapolation on these three datasets, as follows.

The extrapolation experiments are performed on all scenes of the \textbf{7-scenes dataset}. For each scene in the 7-scenes dataset, there are multiple reference sequences, thus we take one of the reference traverses/sequences as our anchor reference trajectory. We then discard the remaining reference sequences\footnote{If we do not discard other reference sequences during extrapolation experiment, they overlap with target extrapolated/regressed descriptors and make the experimental setup less challenging.} to get the original sparse map $M_{sparse}$. Then on the selected reference sequence, we select every $K$th sample (where $K=50$) as our anchor point. Then for each anchor point, we sample target points uniformly in the $x$ and $y$ direction keeping the viewing direction and $z$ fixed to get the dense extrapolated map $M_{dense}$. The sampling of target points is done with a fixed step size $e_{step}$ and a maximum spatial span $e_{span}$ for extrapolation. We use a step size of $e_{step}=0.05$ meters for all seven scenes and the spatial span $e_{span}$ is set to cover the complete area of the scene. Examples of this extrapolation are shown in Fig.~\ref{Fig:interpolationandextrapolation_examples7scenes} for the 7-scenes dataset.

The \textbf{Synthetic Shop Facade dataset} provides a query sequence, a single anchor reference sequence and multiple target reference points sampled uniformly over a fixed grid across this anchor reference sequence. We use this already provided distinction to get $M_{sparse}$ and $M_{dense}$. The query, anchor and target extrapolated points contain novel viewpoints of the same scene and we refer the reader to the author's~\cite{sattler2019understanding} figure here\footnote{\url{https://github.com/tsattler/understanding_apr}} for visualization of the scene and target point distribution. 

In the case of the \textbf{Station Escalator dataset}, the anchor reference images act as the sparse reference map $M_{sparse}$. Extrapolation on the Station Escalator dataset is straightforward: all images on the reference trajectory act as our anchor points and we regress a target descriptor using each anchor at an offset of 1.8 meters on the x-axis from the anchor reference pose. Then, the target descriptors combined with $M_{sparse}$ descriptors act as our extrapolated map $M_{dense}$.

\subsubsection{Extrapolation methods}
Two descriptor regression methods are compared for extrapolation.
\textbf{Linear Regression (Lin. Reg.)} is the local plane fit method introduced in sub-section~\ref{sec:linearregressionusinglocalplanefit}. For the 7-scenes and the Shop Facade dataset, the O NN anchor points are selected from the reference trajectory, and for the Station Escalator dataset, we select two NN anchor points from each of the two parallel trajectories A and B.

\textbf{Non-linear Regression Network (Non-lin. Reg.)}
is the neural network regression approach from sub-section~\ref{sec:nonlin-reg-network}.

\subsubsection{Extrapolation baselines} 
\label{sec:interpolationexperimentsbaselines}
\textbf{Sparse Map:} The primary baseline for extrapolation is the sparse map $M_{sparse}$, where feature descriptors are only available at sparse poses $P$.

\textbf{3D model:} As mentioned in sub-section~\ref{sec:experimentalsetup}, the Shop Facade dataset already provides distinct anchor reference points and target extrapolation points. Since the images for these target extrapolation points are already available, their corresponding feature descriptors at all poses in the extrapolated map can also be computed. We refer to this method as \textit{3D Model} in our results, where the feature descriptors  at all locations (anchor and non-anchor) in $M_{dense}$ are computed using $G_{distance}$ and no descriptor is regressed. This baseline of \cite{sattler2019understanding} helps us to understand how well our extrapolation performs in comparison to having the ground-truth images at all locations in the extrapolated map. 

\textbf{Oracle retrieval:}
We also show the minimum possible translation error and the corresponding rotation error obtained by an oracle retrieval method, which always retrieves the ground-truth 3D Euclidean closest match in the extrapolated map $M_{dense}$.
These errors indicate the VPR base errors for the used queries, and would only be zero if the query poses coincide with the reference poses in the map.

\begin{table*}[!htb]
    \caption{The extrapolation experiments on the 7-scenes dataset.
    The MTE and MRE are reported.
    The oracle retrieval shows the minimum achievable MTE and the corresponding MRE.
    Best in bold.
      } 
    \centering
    \begin{tabular}{c|c|c|c|c|c|c|c|c|c|c|c}
    \hline
    Metric & Map & Densification & Retrieval & Chess & Fire & Heads & Office & Pumpkin & Redkitchen & Stairs & Avg.\\
    \hline
$MTE$ (m)& $M_{dense}$ & - & \textit{Oracle} &0.083 &0.070 &0.030 &0.072 &0.077 &0.216 &0.119 &0.095\\
\hline
$MTE$ (m) & $M_{sparse}$ & - & VPR  &0.318 &0.348 &\textbf{0.158} &0.383 &0.589 &0.567 &0.600 &0.423\\
$MTE$ (m) & $M_{dense}$ & \textit{Lin. Reg.} & VPR &0.245 &0.310 &0.163 &0.338 &0.426 &0.444 &0.532 &0.351\\
$MTE$ (m) & $M_{dense}$ & \textit{Non-lin. Reg.} & VPR &\textbf{0.167} &\textbf{0.279} &0.159 &\textbf{0.264} &\textbf{0.346} &\textbf{0.427} &\textbf{0.430} &\textbf{0.296}\\
\hline
\hline
$MRE$ (\degree) & $M_{dense}$ & - & \textit{Oracle} &28.44 &25.56 &21.25 &58.37 &56.33 &35.97 &23.85 &35.68\\
\hline
$MRE$ (\degree) & $M_{sparse}$ & - & VPR &\textbf{22.54} &20.88 &\textbf{16.49} &\textbf{38.89} &\textbf{44.89} &34.65 &24.32 &\textbf{28.95} \\
$MRE$ (\degree)& $M_{dense}$ & \textit{Lin. Reg.} & VPR &29.04 &\textbf{18.49} &16.62 &39.10 &61.96 &\textbf{33.00} &25.41 &31.95\\
$MRE$ (\degree)& $M_{dense}$ & \textit{Non-lin. Reg.} & VPR &26.87 &22.02 &16.54 &47.95 &58.90 &36.33 &\textbf{21.29} &32.84\\
\hline
 \end{tabular}
    \label{tab:resultsextrapolation7scenes}
\end{table*}

\begin{figure*}[htbp]
\begin{center}
\includegraphics[width=1.0\linewidth]{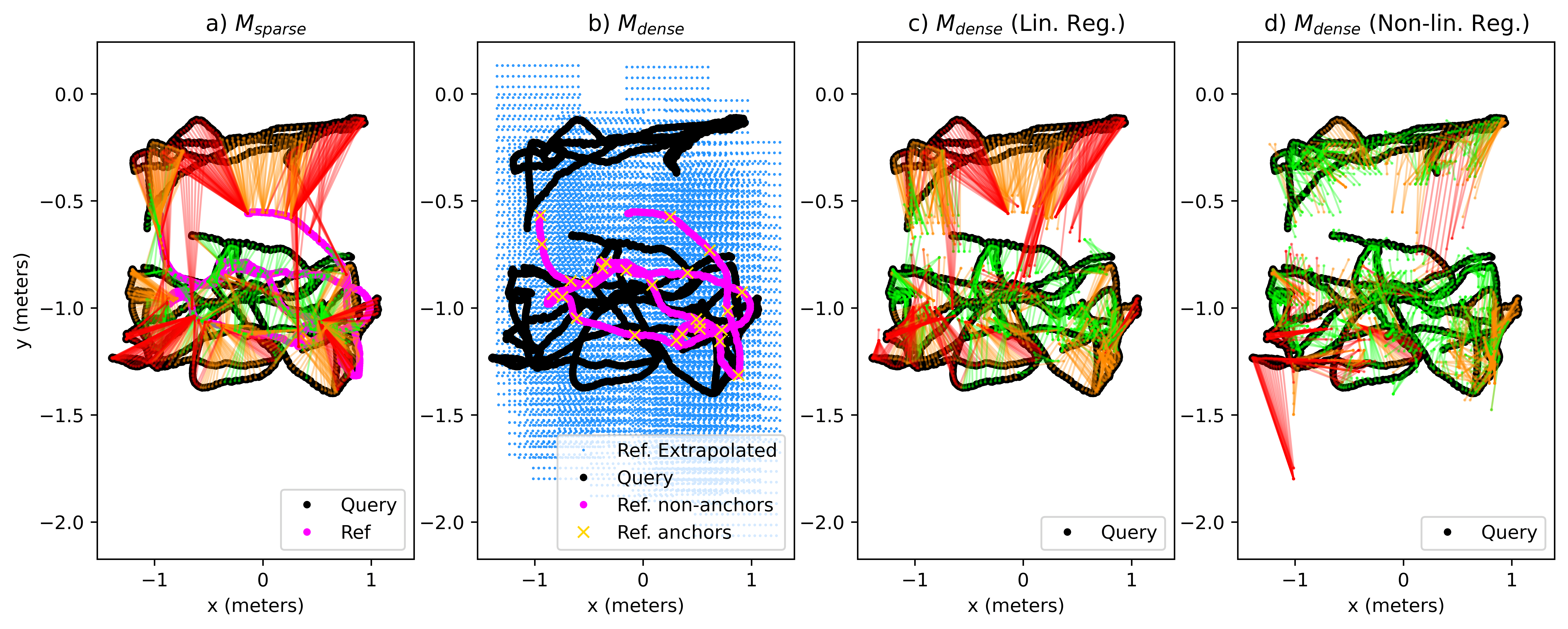}
\end{center}
\caption{Extrapolation experiments on the Office scene of the 7-scenes dataset. (a) The matches between the query and the reference points for the sparse map $M_{sparse}$, (b) the poses in the densified map $M_{dense}$, (c) the matches in map densified using \textit{Lin. Reg.}, (d) the matches in map densified using \textit{Non-lin. Reg.}.  All matches are color-coded as \textit{green}, \textit{orange}, and \textit{red} with increasing 3D Euclidean distance in the physical space. The reference poses in (c) and (d) are the same as in (b) and thus are not shown to avoid cluttering. The non-linearly densified map (d) clearly leads to better performance than other maps, albeit with some failure cases towards the bottom-left of the plot.}
\label{Fig:office_extrapolation}
\end{figure*}  

\subsubsection{Extrapolation results}
We report the extrapolation results in Table~\ref{tab:resultsextrapolation7scenes} for the originally sparse, linearly extrapolated, and non-linearly extrapolated maps for all the seven scenes in the 7-scenes dataset. The matches between the query and the reference trajectories for the extrapolation experiment are shown in Fig.~\ref{Fig:office_extrapolation} for the Stairs scene of the 7-scenes dataset as an example. It can be seen that extrapolation leads to significant performance improvement over no extrapolation in terms of translation error. By using extrapolation we match descriptors closer to the query trajectory. We also note that the non-linear regression model $H$ performs better than the linear regression model, indicating that extrapolating across the trajectory requires a non-linear approach to handle the complexity of the feature space. We do not see performance improvement in translation error due to extrapolation on the Heads scene, where the query and the reference trajectories are already relatively close to each other compared to the other scenes. Moreover, we observe that with the current map densification setup, we cannot improve angular estimation.
However, it is important to notice that even retrieving the Euclidean closest match in physical space leads to an increase in rotation error, as shown by \textit{Oracle} retrieval in Tables~\ref{tab:resultsextrapolation7scenes} and~\ref{tab:resultsextrapolationshopfacade}. We further discuss this increase in rotation error and the reasons behind it in Section~\ref{sec:discussion}.  

The same findings are extended to the Synthetic Shop Facade dataset as reported in Table~\ref{tab:resultsextrapolationshopfacade}. We see performance improvement thanks to extrapolation and the non-linear regression model $H$ outperforms linear regression. We also observe that the VPR performance of the non-linearly extrapolated map (\textit{Non-lin. Reg.}) is similar to the map densified using 3D modelling, which suggests that the trained non-linear regression model $H$ closely regresses the original descriptors, without access to the images at the target poses. 

\begin{figure*}[htbp]
\begin{center}
\includegraphics[width=1.0\linewidth]{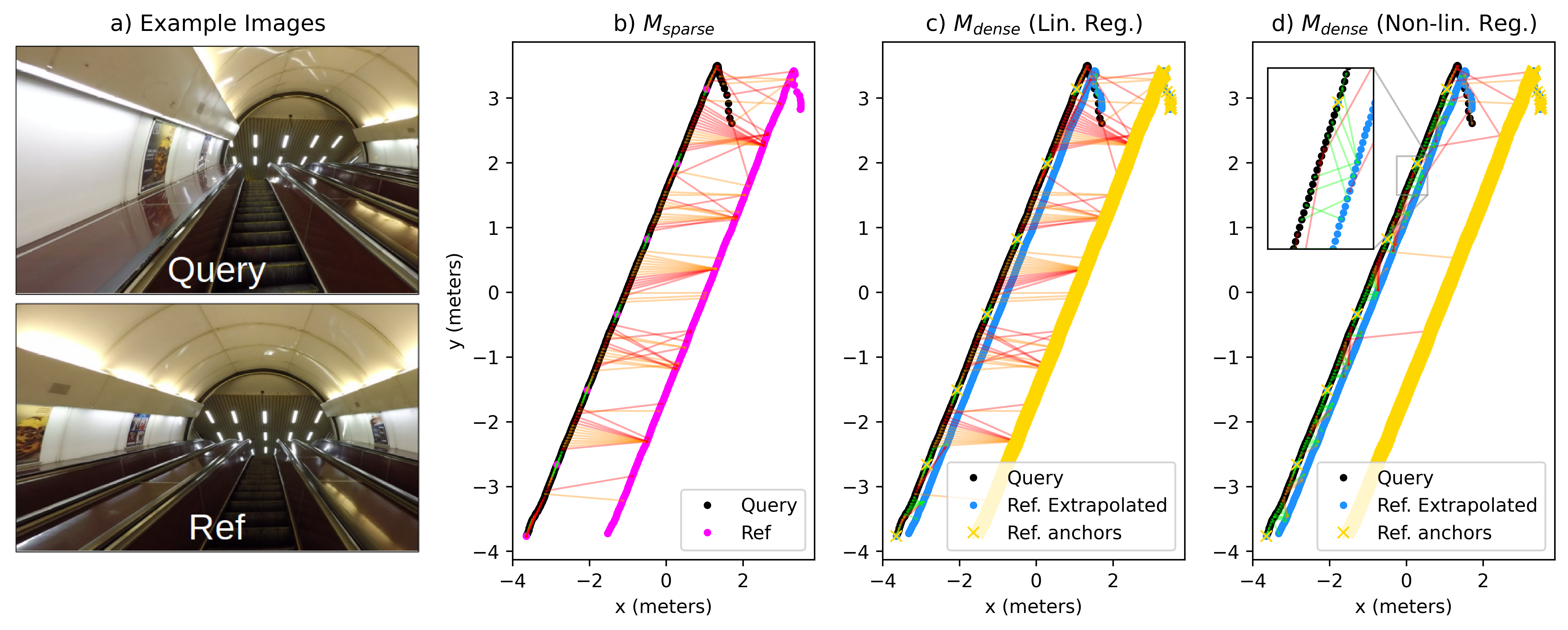}
\end{center}
\caption{Extrapolation experiments on the Station Escalator dataset. (a) Exemplar query and reference images. Then the matches between the query and the reference points for the (b) original sparse map $M_{sparse}$, (c) linearly regressed (\textit{Lin. Reg.}) map $M_{dense}$ and (d) non-linearly regressed (\textit{Non-lin. Reg.}) map $M_{dense}$. These matches are color-coded as \textit{green}, \textit{orange}, and \textit{red} with increasing 3D Euclidean distance in the physical space. Extrapolation with non-linear regression network $H$ is done using only the points on the anchor reference trajectory in \textit{yellow} on the right, whereas the sparse anchor points in \textit{yellow} on the query trajectory are only used at training time.}
\label{Fig:escalator_results}
\end{figure*}

The results on the Station Escalator dataset also support the motivation of this work, since we are able to significantly improve the localization accuracy, as reported in Table~\ref{tab:resultsextrapolationshopfacade}. We also show the qualitative results on the Station Escalator dataset in Fig.~\ref{Fig:escalator_results}. These results highlight the utility of descriptor regression in cases where parallel traverses are common, such as highway lanes, train tracks, escalators, and many such laterally viewpoint-varied paths.

We observe more benefits of non-linear descriptor regression on the Station Escalator dataset than on other datasets.
Linear regression does not work well on this dataset; the selected anchor poses  are too distant from the query trajectory.
Recall that for this dataset the training pairs include sparse samples (every $K$-th image) from both the query and reference traverses
to increase the variance in the training data,
as there are only two traverses in total  in this dataset.
Still, our extrapolation experiments do not extrapolate to the exact query locations but to close-by locations. We observe that training with similar relative pose differences as those observed at test time leads to performance benefits. In a real-world application, if only sparsely sampled images are collected for parallel trajectories, the pose differences are representative to train a regression model and densify the trajectories for improved localization accuracy.

\begin{table*}[htbp]
    \caption{
    The extrapolation experiments on the Synthetic Shop Facade and the Station Escalator datasets.
    The MTE and MRE are reported.
    The oracle retrieval shows the minimum achievable MTE and the corresponding MRE.
    Best in bold.
    }
    \centering
    \begin{tabular}{c||c|c|c|c|c}
    \hline
    Metric & Map & Densification & Retrieval & Shop Facade & Station Escalator\\
    \hline
$MTE$ (m) & $M_{dense}$ & - & \textit{Oracle} &0.188 & 0.26\\
\hline    
$MTE$ (m) & $M_{sparse}$ & - & VPR &0.705 & 2.17\\
$MTE$ (m) & $M_{dense}$ & \textit{3D Model}~\cite{sattler2019understanding} & VPR &\textbf{0.335} & NA\\
$MTE$ (m) & $M_{dense}$ & \textit{Lin. Reg.} & VPR &0.541 & 2.10 \\
$MTE$ (m) & $M_{dense}$ & \textit{Non-lin. Reg.} & VPR &0.344 & \textbf{0.94}\\
\hline
\hline
$MRE$ (\degree)& $M_{dense}$ & - & \textit{Oracle} &11.25 & 9.45\\
\hline
$MRE$ (\degree) & $M_{sparse}$ & - & VPR &\textbf{10.99} & \textbf{8.54} \\
$MRE$ (\degree) &$M_{dense}$ & \textit{3D Model}~\cite{sattler2019understanding} & VPR &11.13 & NA \\
$MRE$ (\degree) &$M_{dense}$ & \textit{Lin. Reg.} & VPR &\textbf{10.99} & 8.60 \\
$MRE$ (\degree) &$M_{dense}$ & \textit{Non-lin. Reg.} & VPR &11.13 & 8.99 \\
\hline
 \end{tabular}
    \label{tab:resultsextrapolationshopfacade}
\end{table*}

\subsection{Interpolation experiments}
\label{sec:interpolationexperiments}
We now explain the setup used for interpolation experiments, followed by the methods and baselines, and then the corresponding results and discussion.

\subsubsection{Interpolation setup}

We perform the interpolation experiments on all the scenes in the 7-scenes dataset. 
Similar to the extrapolation setup, interpolation uses the same concept of a sparse map $M_{sparse}$ and a dense map $M_{dense}$, though for the interpolation experiments these maps are defined differently than for the extrapolation experiments.
For interpolation, 
the full reference trajectory of a scene is used as the \textit{ground-truth} dense map $M_{dense}$.
We then sub-sample the reference trajectories by a factor of $K = 50$, such that the consecutive images in a trajectory still contain visual content overlap.
This reduced set of references is used as the sparse map $M_{sparse}$.
The \textit{ground-truth} dense map serves as a baseline that can assess the performance of VPR if densely sampled reference images would be available, while the sub-sampled version shows the performance when only a sparse set of reference images are available. 
Examples of this sub-sampling are shown in Fig.~\ref{Fig:interpolationandextrapolation_examples7scenes}.
For CoPR, the poses in $P$ from the sparse map act as our anchor poses, while the additional poses $P'$ found in the ground-truth dense map act as the target poses.
All feature descriptors in $M_{sparse}$ and the query descriptors are computed using the feature encoder $G_{distance}$ explained in sub-section~\ref{sec:featureencoder}. 


\subsubsection{Interpolation methods}
The compared descriptor regression methods are
the simple \textbf{Linear Interpolation (Lin. Interp.)} 
from sub-section~\ref{sec:weightedlinearinterpolation}; 
the \textbf{Linear Regression (Lin. Reg.,)} from sub-section~\ref{sec:linearregressionusinglocalplanefit}; 
and the \textbf{Non-linear Regression Network (Non-lin. Reg.)} from sub-section~\ref{sec:nonlin-reg-network}.

\subsubsection{Interpolation baselines}
\textbf{Sparse map:} The primary baseline for interpolation is the sparse map $M_{sparse}$, where feature descriptors are only available at sparse poses $P$.

\textbf{Ground-truth dense map:} Unlike the extrapolation experiments where we do not have true images (and hence descriptors) available at target poses, in the case of interpolation experiments we do have these true images. Thus, this \textit{Ground-Truth} (GT) dense map $M_{dense}$ is a baseline that serves the true descriptors for the target poses.

\textbf{Oracle retrieval:} We also show again the minimum possible translation error and the corresponding rotation error from the oracle retrieval method,
as defined in sub-section~\ref{sec:interpolationexperimentsbaselines}.

\subsubsection{Interpolation results}
\begin{figure*}[htbp]
\begin{center}
\includegraphics[width=1.0\linewidth]{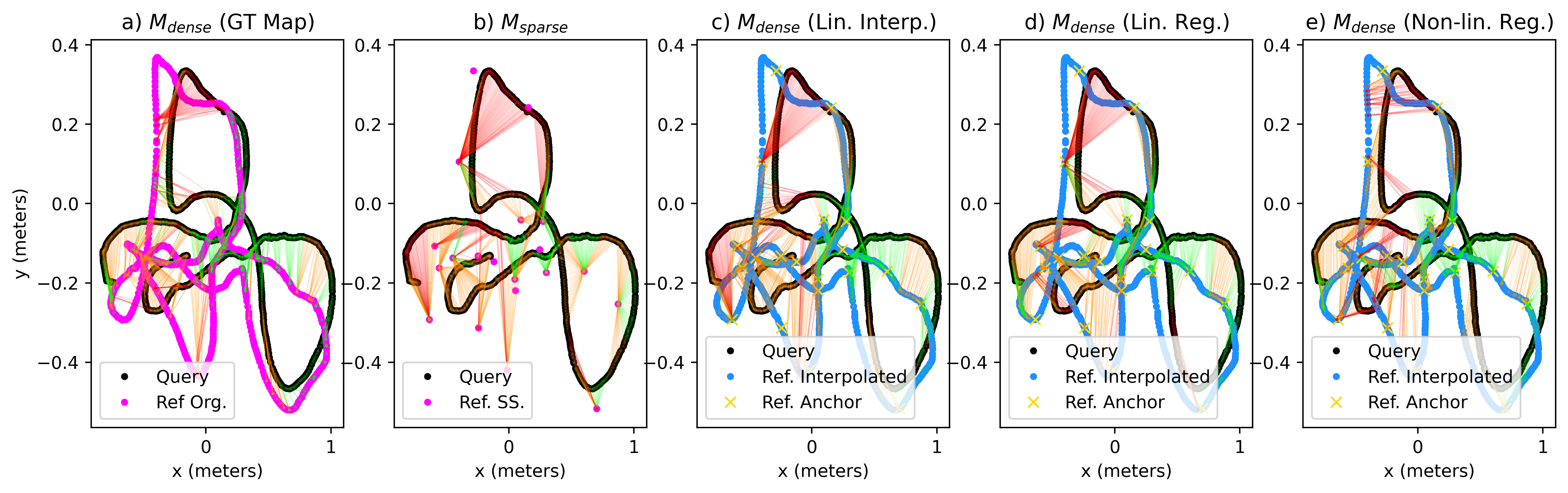}
\end{center}
\caption{Interpolation experiments on the Heads scene of the 7-scenes dataset. The matches between the query and the reference trajectories in (a) the \textit{GT} dense map $M_{dense}$, (b) the sparse map $M_{sparse}$, (c) the linearly regressed (\textit{Lin. Reg.}) map $M_{dense}$, (d) the linearly interpolated (\textit{Lin. Interp.}) map $M_{dense}$ and (e) the non-linearly regressed (\textit{Non-lin. Reg.}) map $M_{dense}$ given K=50. The matches are color-coded as \textit{green}, \textit{orange}, and \textit{red} with increasing 3D Euclidean distance in the physical space.}
\label{Fig:heads_interpolation}
\end{figure*}

The results for all the methods and baselines for the interpolation experiment on the 7-scenes dataset are reported in Table~\ref{tab:resultsinterpolation} for all the seven scenes. The VPR matches between the query and reference trajectories for the Heads scene are shown in Fig.~\ref{Fig:heads_interpolation}. 
We can see a general decrease in localization error when moving from the sparse map $M_{sparse}$ to the \textit{GT} dense map $M_{dense}$. 
Interestingly, we also see that even simple linear regression (\textit{Lin. Reg.} and \textit{Lin. Interp.}) can solve this problem well and is often the best-performing technique. 
Note though that linear regression is done using multiple anchor points which constraints the problem setting, while the non-linear regression network $H$ only uses one anchor point. Nevertheless, this experiment shows that map densification even via interpolating along the trajectory is helpful, although has lesser benefits than extrapolation across the trajectory. 

We will discuss the observed differences between the interpolation and extrapolation experiments in more detail in the Discussion,
Section~\ref{sec:discussion}.

\begin{table*}[htbp]
    \caption{The interpolation experiments for the 7-scenes dataset at K=50.
    The MTE and MRE are reported.
    The oracle retrieval shows the minimum achievable MTE and the corresponding MRE.
    Best is in Bold. 
    } 
    \centering
    \begin{tabular}{c||c|c|c|c|c|c|c|c|c|c|c}
    \hline
    Metric & Map & Densification & Retrieval & Chess & Fire & Heads & Office & Pumpkin & Redkitchen & Stairs & Avg. \\
    \hline
$MTE$ (m) & $M_{dense}$ & - & \textit{Oracle} &0.109 &0.183 &0.097 &0.117 &0.115 &0.129 &0.132 &0.126\\ 
$MTE$ (m) & $M_{dense}$ & \textit{GT Map} & VPR &0.165 &0.255 &0.158 &0.207 &0.242 &0.219 &0.261 &0.215\\
\hline
$MTE$ (m) & $M_{sparse}$ & - & VPR &0.210 &0.322 &0.212 &0.237 &0.250 &0.271 &0.263 &0.252 \\
$MTE$ (m) & $M_{dense}$ & \textit{Lin. Interp.} & VPR &0.170 &0.277 &0.202 &\textbf{0.211} &0.257 &\textbf{0.220} &\textbf{0.257} &0.227 \\
$MTE$ (m) & $M_{dense}$ & \textit{Lin. Reg.} & VPR &\textbf{0.169} &\textbf{0.257} &\textbf{0.165} &0.216 &\textbf{0.214} &0.224 &0.262 & \textbf{0.215}  \\
$MTE$ (m) & $M_{dense}$ & \textit{Non-lin. Reg.} & VPR &0.178 &0.264 &0.184 &0.221 &0.259 &0.260 &0.278 &0.234 \\
\hline
\hline
$MRE$ (\degree) & $M_{dense}$ & - & \textit{Oracle} &22.81 &26.65 &20.91 &43.56 &37.58 &31.31 &29.71 &30.36 \\ 
$MRE$ (\degree) & $M_{dense}$ & \textit{GT Map} & VPR &17.69 &19.71 &16.49 &32.13 &36.24 &22.49 &19.55 &23.47 \\
\hline
$MRE$ (\degree) & $M_{sparse}$ & - & VPR &20.75 &\textbf{19.15} &19.34 &35.01 &\textbf{33.96} &27.27 &\textbf{19.16} &24.94 \\
$MRE$ (\degree) & $M_{dense}$ & \textit{Lin. Interp.} & VPR &21.73 &20.65 &17.93 &\textbf{34.65} &39.15 &26.27 &19.92 & 25.75 \\
$MRE$ (\degree) & $M_{dense}$ & \textit{Lin. Reg.} & VPR &\textbf{20.09} &20.00 &\textbf{17.20} &35.63 &34.00 &\textbf{24.16} &19.72 &\textbf{24.40} \\
$MRE$ (\degree) & $M_{dense}$ & \textit{Non-lin. Reg.} & VPR &22.09 &19.45 &20.13 &39.73 &39.03 &27.17 &20.25 &26.83 \\
\hline
 \end{tabular}
    \label{tab:resultsinterpolation}
\end{table*}

\subsection{Map densification with different feature encoders}
\label{sec:MapdensificationwithDifferentFeatureEncoders}
Next, we test that using the non-linear regression model $H$ for extrapolating across anchor points is beneficial for all discussed feature encoders. This is reported in Table~\ref{tab:MTEofDifferentEncoderswithandwithoutCoPR}. However, the corresponding localization accuracy is limited by the localization performance of the respective feature encoder. The MTE is reported for all three types of feature encoders on the sparse map $M_{sparse}$ and the non-linearly regressed (\textit{Non-lin. Reg.}) map $M_{dense}$ for the 7-scenes dataset and the Synthetic Shop Facade dataset. Such a generic boost of performance using map densification supports that CoPR can utilize inherent benefits of different types of feature encoders, for example, the domain generalization of $G_{triplet}$ and $G_{relative}$, and the viewpoint variance of $G_{distance}$.

\begin{table*}[htbp]
    \caption{
    The effect of CoPR on different feature encoders on all scenes from the 7-scenes dataset, and on the Synthetic Shop Facade dataset.
    In the case of $G_{triplet}$, we use the triplet loss as motivated by the authors~\cite{arandjelovic2016netvlad} but do not use the VLAD descriptor module to keep the backbone the same for a fair comparison across all encoders.
    } 
    \centering
    \begin{tabular}{|c|c|c|c|c|c|c|}
    \hline
    \textbf{Feature Encoder} &  \multicolumn{2}{|c|}{$G_{triplet}$ } &  \multicolumn{2}{|c|}{$G_{relative}$ } & \multicolumn{2}{|c|}{$G_{distance}$} \\
    \hline
    \textbf{Reference Map} & $M_{sparse}$ & $M_{dense}$ & $M_{sparse}$ & $M_{dense}$ & $M_{sparse}$ & $M{dense}$ \\
    \hline
7-scenes - Chess &0.600 &\textbf{0.450} &0.379 &\textbf{0.260} &0.318 &\textbf{0.167} \\
7-scenes - Fire &0.612 &\textbf{0.542} &0.414 &\textbf{0.296} &0.348 &\textbf{0.279} \\
7-scenes - Heads &0.227 &\textbf{0.215} &0.166 &\textbf{0.147} &\textbf{0.158} &0.159 \\
7-scenes - Office &0.680 &\textbf{0.589} &0.455 &\textbf{0.246} &0.383 &\textbf{0.264} \\
7-scenes - Pumpkin &1.306 &\textbf{1.208} &0.720 &\textbf{0.479} &0.589 &\textbf{0.346} \\
7-scenes - Redkitchen &0.960 &\textbf{0.783} &0.691 &\textbf{0.451} &0.567 &\textbf{0.427} \\
7-scenes - Stairs &\textbf{0.699} &0.780 &0.673 &\textbf{0.374} &0.600 &\textbf{0.430} \\
\hline
Synthetic Shop Facade &2.234 &\textbf{1.419} &2.219 &\textbf{1.641} &0.705 &\textbf{0.344} \\
\hline
Average &0.915 &\textbf{0.748} &0.715 &\textbf{0.487} &0.458 &\textbf{0.302} \\
\hline
 \end{tabular}
    \label{tab:MTEofDifferentEncoderswithandwithoutCoPR}
\end{table*}

\subsection{Map-density vs localization accuracy}
\begin{figure}[htbp]
\begin{center}
\includegraphics[width=1.0\linewidth]{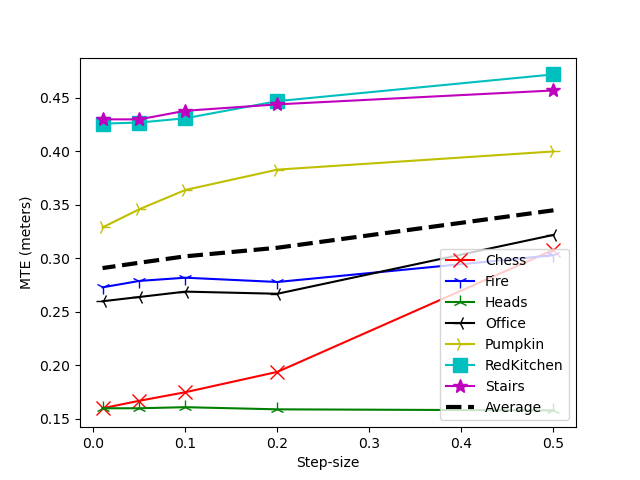}
\end{center}
\caption{The increase in MTE by increasing the step size $e_{step}$ for all scenes of the 7-scenes dataset. A larger step size leads to sparser maps which increases the translation error, while a smaller step size leads to denser maps which are useful for accurate localization.}
\label{Fig:stepsizevslocalizationaccuracy}
\end{figure}

The motivation presented in this work suggests that the denser the reference map, the lesser will be the localization error of a VPR-based localization system. In our work, this map density is modelled with the step size $e_{step}$. Therefore, in this sub-section, we show the effect of increasing map density on the localization error by using extrapolation with non-linear regression model $H$ and feature encoder $G_{distance}$ for the 7-scenes dataset. This direct relation between the step size $e_{step}$ and the MTE is presented in Fig.~\ref{Fig:stepsizevslocalizationaccuracy}. Decreasing the step size leads to denser extrapolated maps which then leads to a decrease in MTE for the non-linearly extrapolated (\textit{Non. Lin. Reg.}) map $M_{dense}$. The performance benefits for the scenes depend on the underlying scene geometry and the quality of descriptor regression. For example, in the case of Heads scene, the query poses and the sparse reference poses in $M_{sparse}$ are already close to each other, thus we do not see any performance benefits due to densification. While in other scenes, we see that map densification is helpful and is related to the level of map densification modelled with the step size $e_{step}$.

\subsection{Benefits of CoPR for RPE}
\label{sec:DescriptorRegressionCanBenefitRelativePoseEstimation}
In this section, we look into the relation of CoPR with RPE and hence CtF localization, as discussed in sub-section~\ref{sec:relatingcoprwithctf}. In this experiment, we make this argument concrete by illustrating that situations exist where a sparse map leads to incorrect coarse retrieval of a visually similar image descriptor taken at an arbitrarily far location, which can in turn lead to the failure of CtF approaches. We argue that this error source is fundamentally due to the retrieval step, not due to the subsequent RPE step, and demonstrate that map densification could tackle this error source in some cases.

We create exemplar cases in the 7-scenes dataset where such an effect can be easily observed. A reference database of four sparsely sampled reference images around a query image is created for a given scene and a fifth \textit{stray} reference image is added to this reference database. This stray image is taken from a completely different scene that has no real physical overlap with the query image. We use the feature encoder $G_{relative}$ for image-retrieval and the non-linear descriptor regression network $H$ to regress the expected descriptor at the query location given the nearest anchor reference descriptor. This regressed descriptor acts as the descriptor for a hypothetical 6th image in the reference database at the query location. 

\begin{figure*}[htbp]
\begin{center}
\includegraphics[width=1.0\linewidth]{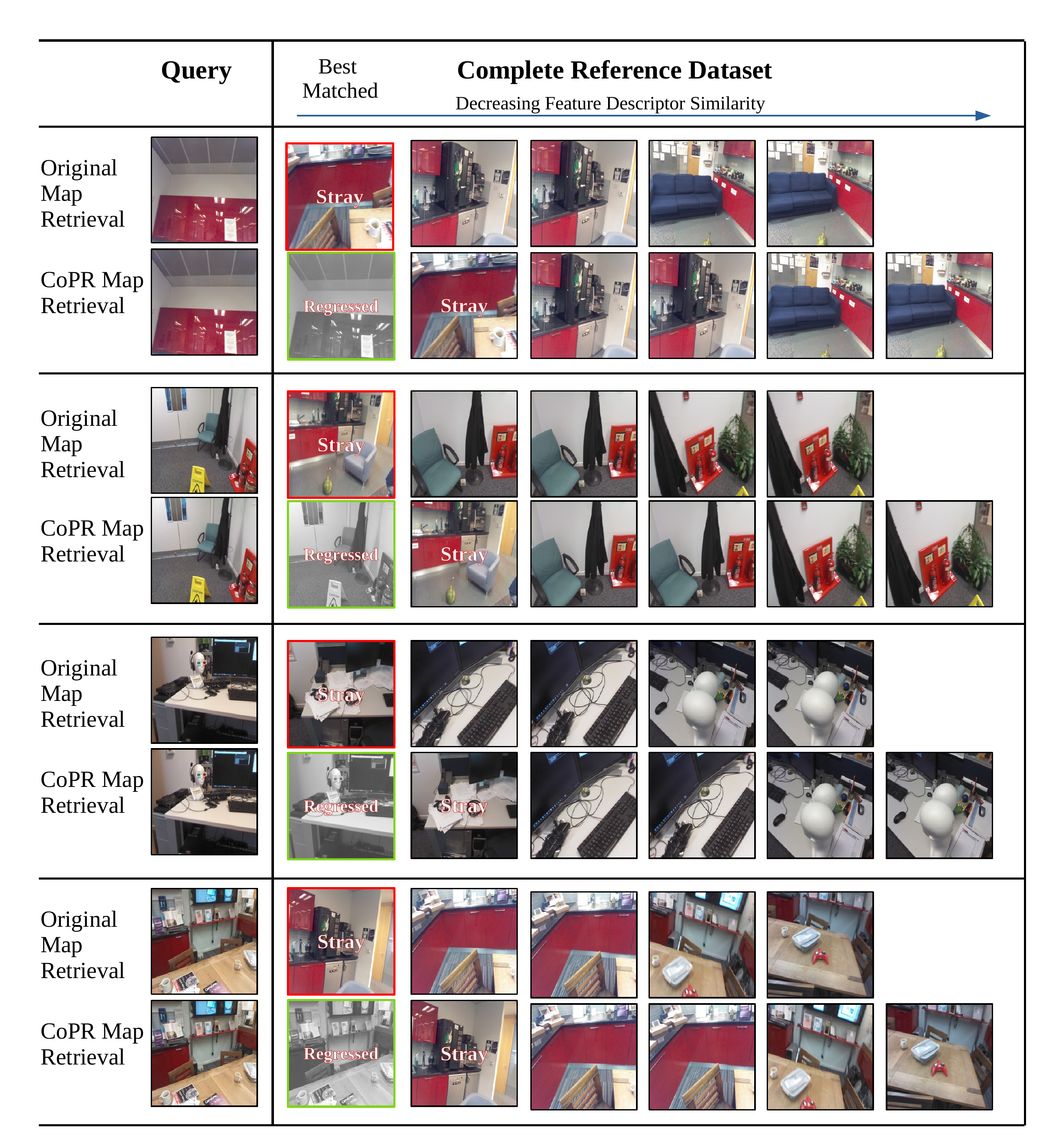}
\end{center}
\caption{The exemplar cases where image-retrieval fails to retrieve useful coarse estimates for RPE in a sparse reference map. By regressing the expected descriptor at the query pose, we show that map densification could lead to robustness against such failure cases. The grayscale image in the reference set is only added for the reader's reference and represents only a hypothetical image for the regressed descriptor at the query pose since we do not synthesize images but only regress image descriptors. The \textit{green} bounding box represents a correct match and the \textit{red} bounding box represents an incorrect match.}
\label{Fig:relposefailure}
\end{figure*}

The objective of this experiment is to show that in the absence of the regressed descriptor, the stray image is selected as the best match for the query image, while in the presence of the regressed descriptor, the stray image is pushed downwards in the list of retrieved images ranked by their matching scores. Please note that in the case where the stray image is chosen as the best match, the localization error can be arbitrarily large, as a different scene can be quite far. We show four such example cases in Fig.~\ref{Fig:relposefailure} from the 7-scenes dataset, where we can observe that in the absence of the regressed descriptor, the stray image is chosen as the best match by the image-retrieval system. Since such stray cases are shown to exist in multiple scenes of the 7-scenes dataset, which is a small-scale dataset, this effect would amplify even further in spatially larger scenes due to the increased chances of perceptual aliasing.

Thus, without CoPR, sparse reference maps \textit{could} lead to incorrect coarse retrieval, where the coarse pose estimate can be arbitrarily far-away and hence cannot be corrected by CtF approaches.
By using CoPR, 
reference descriptors of the correct scene now appear close to the query descriptor.
Finding all references near the query in the feature space thus identifies similar scenes, allowing to at least represent localization ambiguity and ideally obtain a correct best match.
Without CoPR only the incorrect scene would have matched the query.
Better retrieval also benefits CtF approaches, since the RPE step is only valid if the retrieved reference pose represents the correct scene.
These constructed cases illustrate that CoPR and CtF are complementary approaches to improve VPR-based localization accuracy.
Please note that this analysis does not demonstrate that CoPR prevents false positives as a general rule, but that it is possible to construct cases where the complementarity of CoPR and CtF can be observed. Future works may investigate this further.

\subsection{Computational details}
\label{computational_details}
Finally, we report the sizes of the sparse and dense maps, the time spent $t_{dense}$ on creating the dense maps $M_{dense}$ using $H$, and the training times $t_{train}$ of model $H$ for all the datasets in Table~\ref{tab:CoPR_computationalfootprint}. For the 7-scenes dataset, the results are reported for the Office scene. The retrieval time $t_{retr}$ in VPR is the sum of the time $t_{enc}$ required to encode a query image into a feature descriptor and the time $t_{match}$ spent to find the NN match of this descriptor in the map. Since the encoding time is several times higher than the efficient NN search, the retrieval time is not too affected by map densification. Please note that the timings are not comparable between the datasets due to differences in map content (i.e., descriptors).

\begin{table}[htbp]
    \caption{The computational footprint of CoPR, please see accompanying text for details.}
    \centering
    \begin{tabular}{|c|c|c|c|c|c|}
\hline
\textbf{} & \textbf{Map} &\textbf{7-scenes} &\textbf{Shop Fac.} &\textbf{Stat. Esc.} \\
\hline
\textbf{$t_{train} (sec)$} & - &510 &540 &960 \\
\hline
\textbf{$t_{dense} (msec)$} & - &12.8 &2.241 &0.32 \\
\hline
\textbf{$t_{enc} (msec)$} & - &6.16 &8.39 &5.88 \\
\hline
\multirow{2}{*}{\textbf{$t_{match} (msec)$}} &$M_{sparse}$ &0.02 &0.1 &0.02 \\
&$M_{dense}$ &0.05 &0.32 &0.08 \\
\hline
\multirow{2}{*}{\textbf{$t_{retr} (msec)$}} &$M_{sparse}$ &6.18 &8.49 &5.90 \\
&$M_{dense}$ &6.21 &8.71 &5.96 \\
\hline
\multirow{2}{*}{Map Size (\#)} &$M_{sparse}$ &1000 &231 &337 \\
&$M_{dense}$ &13000 &2531 &667 \\
\hline
 \end{tabular}
    \label{tab:CoPR_computationalfootprint}
\end{table}

\section{Discussion}
\label{sec:discussion}
In this section, we identify the major limitations of our work and areas that need further investigation.

\textbf{Angular error:} In both the interpolation and extrapolation experiments, it is clear that our approach does not improve angular localization accuracy, as reported in Tables~\ref{tab:resultsextrapolation7scenes}, \ref{tab:resultsextrapolationshopfacade} and \ref{tab:resultsinterpolation}. However, it is also important to note that retrieving the ground-truth Euclidean closest match in the physical space also leads to an \textit{increase} in angular error ($MRE$). This is because the nearest match in terms of translation may not have the same 3D orientation. Thus, we attribute the increase in rotation error using CoPR to two reasons: firstly, during interpolation and extrapolation experiments, we do not change the angular pose but only the translation pose, given the anchor points, for the target points, and secondly, the encoder $G_{distance}$ does not optimize for angular localization error in its training objective. Thus, reducing both the translation and angular error requires that the Euclidean closest match in the physical space has the closest angular orientation to the query image. Future works could look into the benefits of using distance+orientation based encoder loss along with map densification in a 6-DoF setting. 

\begin{figure}[htbp]
\begin{center}
\includegraphics[width=0.85\linewidth]{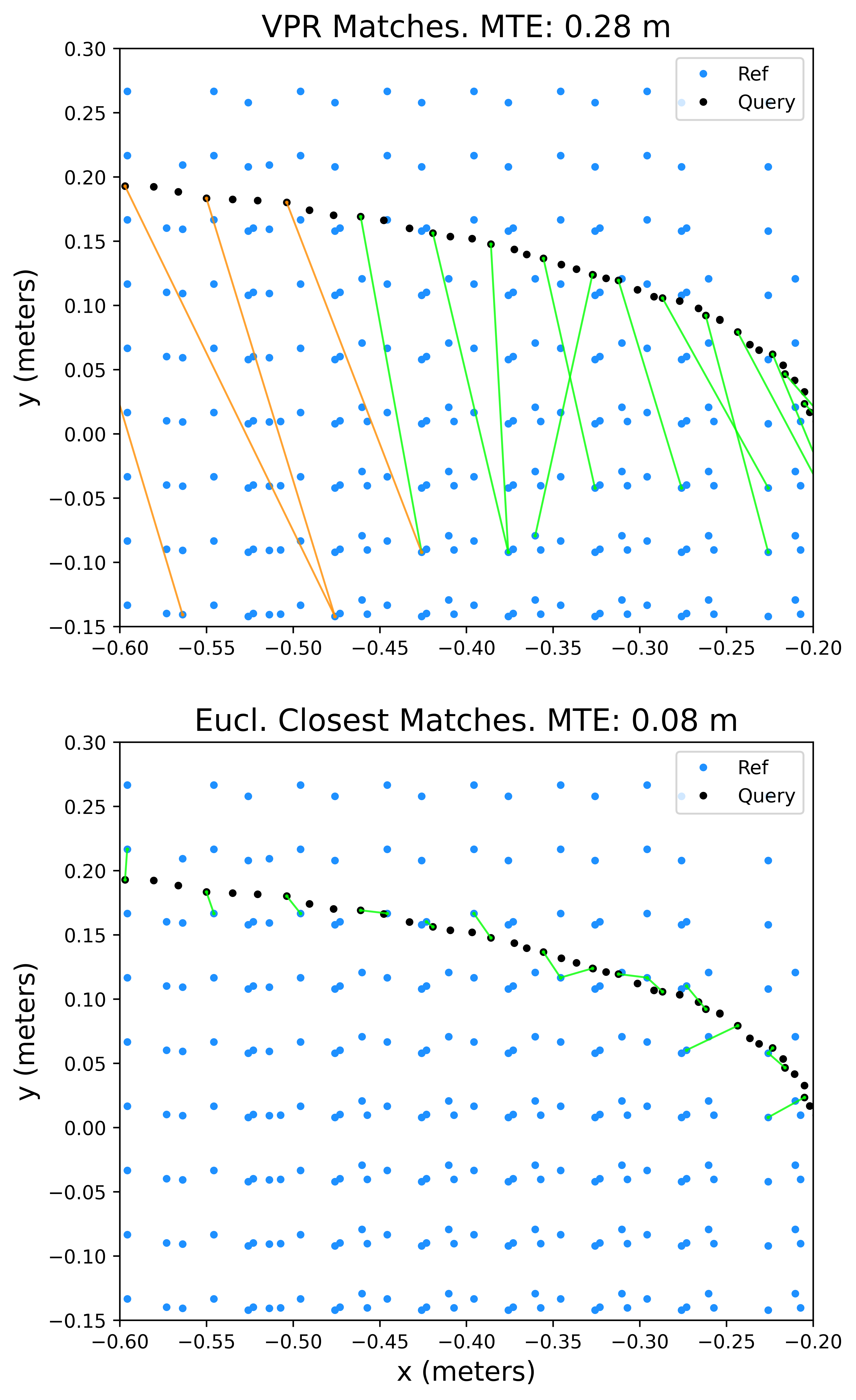}
\end{center}
\caption{The VPR matches (top) and ground-truth 3D Euclidean closest matches in the physical space (bottom) between the query and the reference trajectories in the Fire scene of the 7-scenes dataset for the non-linearly extrapolated (\textit{Non-lin. Reg.}) map $M_{dense}$. The matches are color-coded as \textit{orange} and \textit{green} with increasing 3D Euclidean distance. Although non-linearly regressed target poses (in \textit{blue}) are matched to by VPR, these are not always the Euclidean closest matches in the physical space. Hence there is still room for improvement.}
\label{Fig:extrapolation_vprvsgroundtruth_fire}
\end{figure}

\textbf{Ground-truth closest matches:} Our results on extrapolation show that map densification can lead to a significant decrease in localization error. Moreover, the extrapolation experiments on the Shop Facade dataset also show that the localization performance on the non-linearly extrapolated (Non-lin. Reg.) map $M_{dense}$ is close to the localization performance on a ground-truth (obtained using 3D modelling) dense map $M_{dense}$. However, the localization error given the encoder $G_{distance}$ and the non-linear regression network $H$ is still higher than the minimum possible localization error. We have reported the minimum possible translation error (\textit{Oracle Retrieval}) in $M_{dense}$ in Tables~\ref{tab:resultsextrapolation7scenes}, \ref{tab:resultsextrapolationshopfacade} and \ref{tab:resultsinterpolation}. We further show qualitatively in Fig.~\ref{Fig:extrapolation_vprvsgroundtruth_fire}, the performance that could be achieved by an oracle VPR system that always retrieves the Euclidean closest match in the physical space as the best match in a dense map. This gap in performance presents room for future research in this area. Furthermore, our results only show the generalization of non-linear data-driven regression model $H$ across viewpoints within the same scene, however, generalization across scenes could be the new frontier for CoPR. 

\textbf{Interpolation vs extrapolation}: From our results of the two experiments, it can be noted that the absolute decrease in localization error from interpolation is less than the decrease in localization error from extrapolation.
We hypothesize two reasons for this:
1) the query trajectory has larger relative pose distance to the extrapolated poses than to the interpolated poses,
2) the viewpoint variance vs invariance of VPR encoders (as explained in sub-section~\ref{sec:featureencoder}) acts as a bottleneck, since the VPR system does not necessarily match to the ground-truth Euclidean closest match in the physical space but to \textit{one of the closest} matches.
We expect that major performance benefits, given these experiments, require models that have even better viewpoint variance than the feature encoder $G_{distance}$. This motivates viewpoint-variant VPR for high accuracy, in addition to the existing trends for viewpoint-invariant VPR~\cite{berton2021viewpoint}.

Generally, we find that extrapolation is more useful than interpolation  when a repeated traversal could occur at a laterally offset-ed path. Such trajectories are common to observe in real-world, for example, parallel traverses in outdoor scenes (Shop Facade dataset) and parallel traverses in indoor scenes (Station Escalator dataset). Other examples include lanes on a highway and parallel paths in corridors. However, our results do show that both interpolating and extrapolating descriptors generally give better localization accuracy than using sparser reference maps, which suggests that map densification (CoPR) along the trajectory and/or across the anchor points can be useful for VPR.

\section{Conclusions}
In this paper, we investigated the discrete treatment of places in a VPR map. We have shown that map densification whether using interpolation or extrapolation is helpful to reduce translation error. Our results for the 7-scenes dataset suggest that interpolating along the trajectory is an easier problem and can be solved with simple linear regression in the local neighborhood, however, extrapolation benefits from a non-linear treatment. Moreover, our proposed non-linear regression network only uses a single anchor point for regression, while our linear regression method uses multiple anchor points. We validated that map densification is helpful for feature encoders trained with the three different types of losses, and that the highest accuracy is achieved when using a distance-based loss. Moreover, the benefit of map densification is shown for three datasets: 7-scenes, Synthetic Shop Facade, and Station Escalator, where each of them represents a different type of problem setting. We also discussed that RPE and CoPR address related but complementary problems. We demonstrated through several constructed cases that in a sparse map localization might fail due to perceptual aliasing. RPE cannot recover the true location from a retrieved wrong place. CoPR helps retrieve the correct place, thus solving errors that RPE cannot.

While the distance-based loss function helps to retain viewpoint information among descriptors, we observed that there is still room for improvement in comparison to retrieving the ground-truth Euclidean closest reference descriptors in the physical space. Future works could investigate architectures and loss functions that further enforce the network to learn feature representations useful for retrieving the 3D Euclidean closest match.
As shown in this work, anchor selection and descriptor extrapolation are two separate steps for map densification. In the future a separate treatment of both, i.e., learning good anchors and extrapolating well using multiple anchors, could lead to better map densification. We hope that this work helps to identify the important problem of map densification through Continuous Place-descriptor Regression (CoPR) for VPR and its relation to viewpoint variance, and motivates further research on improving VPR-based localization accuracy through CoPR.

\begin{IEEEbiography}[{\includegraphics[width=1in,height=1.25in,clip,keepaspectratio]{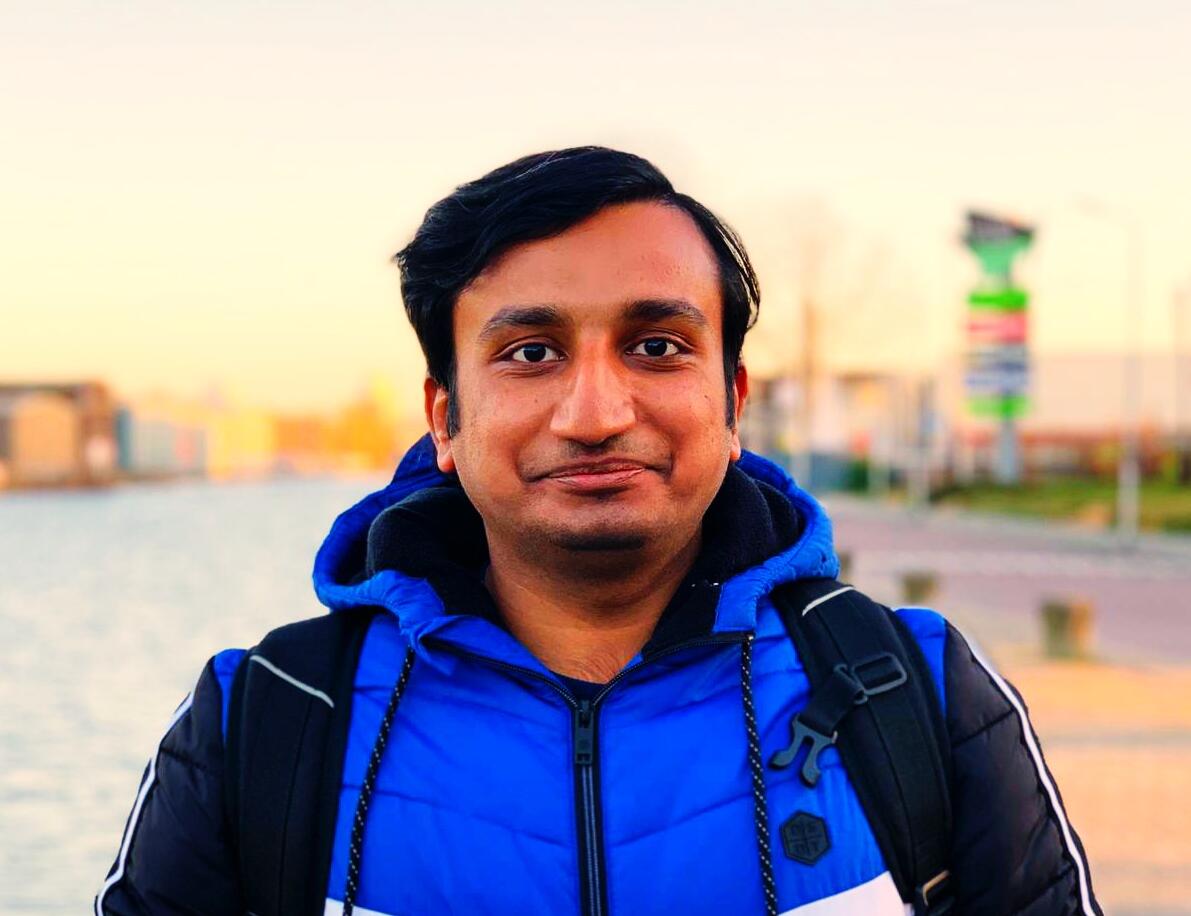}}]{Mubariz Zaffar} is a Ph.D. candidate supervised by Dr. Julian Kooij and Dr. Liangliang Nan in the 3D Urban Understanding (3DUU) Lab at the Delft University of Technology (TUD), and member of the TUD Intelligent Vehicles Group headed by Prof. Dr. Dariu M. Gavrila. He obtained his Master of Science by Dissertation degree in 2020 from the University of Essex and his Bachelor of Electrical Engineering from the National University of Sciences and Technology, Pakistan in 2016. His research interests include place recognition, visual localization and mapping, and representation learning.
\end{IEEEbiography}

\begin{IEEEbiography}[{\includegraphics[width=1in,height=1.25in,clip,keepaspectratio]{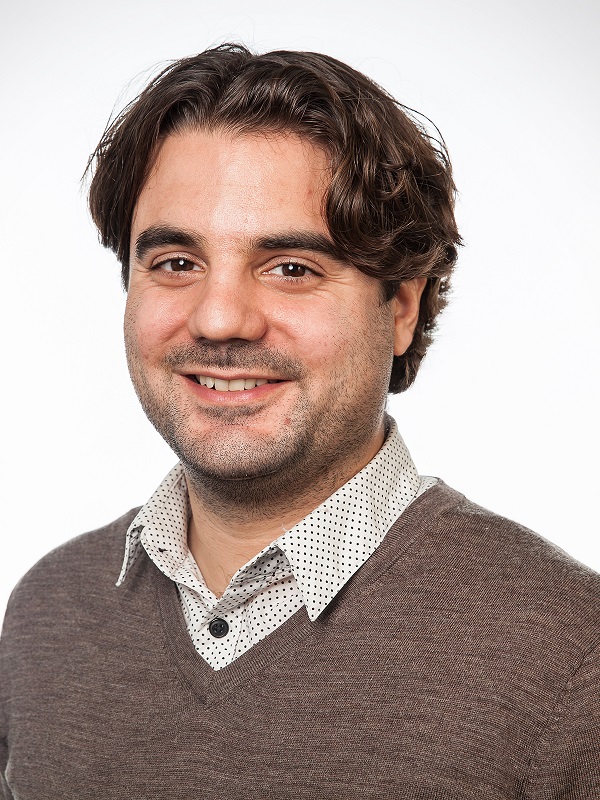}}]{Julian F. P. Kooij}obtained the Ph.D. degree in 2015 at the University of Amsterdam
on visual detection and path prediction for vulnerable road users.
Afterwards, he joined Delft University of Technology, first in the Computer Vision lab
and since 2016 in the Intelligent Vehicles group
where he is currently an Associate Professor.
His research interests include deep representation learning and probabilistic models for multi-sensor localisation, object detection, and forecasting of urban traffic.
\end{IEEEbiography}

\begin{IEEEbiography}[{\includegraphics[width=1in,height=1.25in,clip,keepaspectratio]{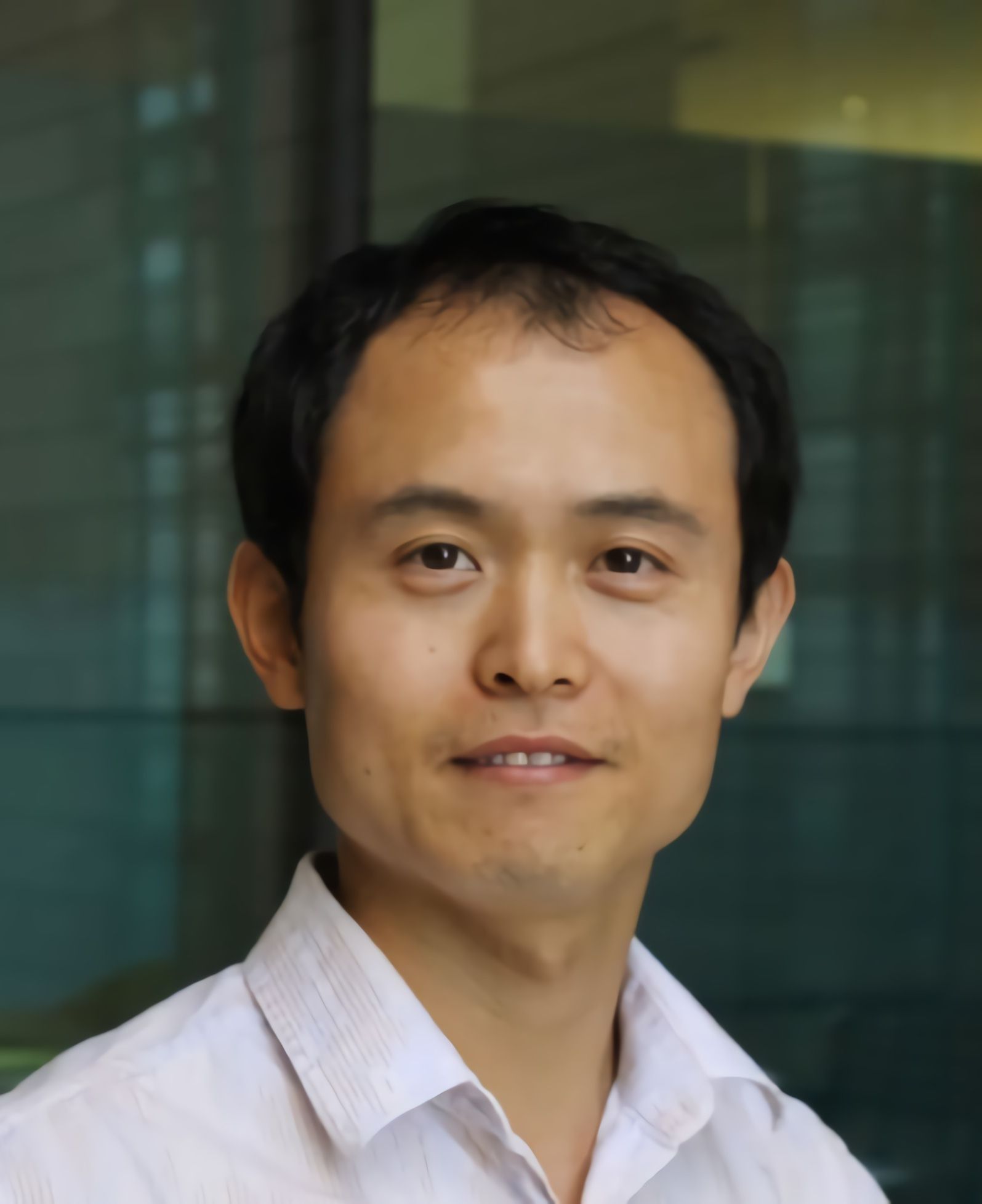}}]{Liangliang Nan} received his Ph.D. degree in mechatronics engineering from the Graduate University of the Chinese Academy of Sciences, China, in 2009. Before joining the Delft University of Technology as an assistant professor in 2018, he worked as a research scientist at the Visual Computing Center, at King Abdullah University of Science and Technology. His research interests are in computer vision, computer graphics, 3D geoinformation, and machine learning.
\end{IEEEbiography}
{
\small
\bibliographystyle{IEEEtran}
\bibliography{root}

\begin{thebibliography}{10}
\providecommand{\url}[1]{#1}
\csname url@samestyle\endcsname
\providecommand{\newblock}{\relax}
\providecommand{\bibinfo}[2]{#2}
\providecommand{\BIBentrySTDinterwordspacing}{\spaceskip=0pt\relax}
\providecommand{\BIBentryALTinterwordstretchfactor}{4}
\providecommand{\BIBentryALTinterwordspacing}{\spaceskip=\fontdimen2\font plus
\BIBentryALTinterwordstretchfactor\fontdimen3\font minus
  \fontdimen4\font\relax}
\providecommand{\BIBforeignlanguage}[2]{{%
\expandafter\ifx\csname l@#1\endcsname\relax
\typeout{** WARNING: IEEEtran.bst: No hyphenation pattern has been}%
\typeout{** loaded for the language `#1'. Using the pattern for}%
\typeout{** the default language instead.}%
\else
\language=\csname l@#1\endcsname
\fi
#2}}
\providecommand{\BIBdecl}{\relax}
\BIBdecl

\bibitem{toft2020long}
C.~Toft, W.~Maddern, A.~Torii, L.~Hammarstrand, E.~Stenborg, D.~Safari,
  M.~Okutomi, M.~Pollefeys, J.~Sivic, T.~Pajdla \emph{et~al.}, ``Long-term
  visual localization revisited,'' \emph{IEEE Transactions on Pattern Analysis
  and Machine Intelligence}, 2020.

\bibitem{piasco2018survey}
N.~Piasco, D.~Sidib{\'e}, C.~Demonceaux, and V.~Gouet-Brunet, ``A survey on
  visual-based localization: On the benefit of heterogeneous data,''
  \emph{Pattern Recognition}, vol.~74, pp. 90--109, 2018.

\bibitem{arandjelovic2016netvlad}
R.~Arandjelovic, P.~Gronat, A.~Torii, T.~Pajdla, and J.~Sivic, ``Netvlad: Cnn
  architecture for weakly supervised place recognition,'' in \emph{Proceedings
  of the IEEE conference on computer vision and pattern recognition}, 2016, pp.
  5297--5307.

\bibitem{kendall2015posenet}
A.~Kendall, M.~Grimes, and R.~Cipolla, ``Posenet: A convolutional network for
  real-time 6-dof camera relocalization,'' in \emph{Proceedings of the IEEE
  international conference on computer vision}, 2015, pp. 2938--2946.

\bibitem{laskar2017camera}
Z.~Laskar, I.~Melekhov, S.~Kalia, and J.~Kannala, ``Camera relocalization by
  computing pairwise relative poses using convolutional neural network,'' in
  \emph{Proceedings of the IEEE International Conference on Computer Vision
  Workshops}, 2017, pp. 929--938.

\bibitem{lowry2015visual}
S.~Lowry, N.~S{\"u}nderhauf, P.~Newman, J.~J. Leonard, D.~Cox, P.~Corke, and
  M.~J. Milford, ``Visual place recognition: A survey,'' \emph{IEEE
  Transactions on Robotics}, vol.~32, no.~1, pp. 1--19, 2015.

\bibitem{zaffar2021vpr}
M.~Zaffar, S.~Garg, M.~Milford, J.~Kooij, D.~Flynn, K.~McDonald-Maier, and
  S.~Ehsan, ``Vpr-bench: An open-source visual place recognition evaluation
  framework with quantifiable viewpoint and appearance change,''
  \emph{International Journal of Computer Vision}, vol. 129, no.~7, pp.
  2136--2174, 2021.

\bibitem{garg2021your}
S.~Garg, T.~Fischer, and M.~Milford, ``Where is your place, visual place
  recognition?'' \emph{arXiv preprint arXiv:2103.06443}, 2021.

\bibitem{sattler2019understanding}
T.~Sattler, Q.~Zhou, M.~Pollefeys, and L.~Leal-Taixe, ``Understanding the
  limitations of cnn-based absolute camera pose regression,'' in
  \emph{Proceedings of the IEEE/CVF conference on computer vision and pattern
  recognition}, 2019, pp. 3302--3312.

\bibitem{balntas2018relocnet}
V.~Balntas, S.~Li, and V.~Prisacariu, ``Relocnet: Continuous metric learning
  relocalisation using neural nets,'' in \emph{Proceedings of the European
  Conference on Computer Vision (ECCV)}, 2018, pp. 751--767.

\bibitem{ding2019camnet}
M.~Ding, Z.~Wang, J.~Sun, J.~Shi, and P.~Luo, ``Camnet: Coarse-to-fine
  retrieval for camera re-localization,'' in \emph{Proceedings of the IEEE/CVF
  International Conference on Computer Vision}, 2019, pp. 2871--2880.

\bibitem{sattler2018benchmarking}
T.~Sattler, W.~Maddern, C.~Toft, A.~Torii, L.~Hammarstrand, E.~Stenborg,
  D.~Safari, M.~Okutomi, M.~Pollefeys, J.~Sivic \emph{et~al.}, ``Benchmarking
  6dof outdoor visual localization in changing conditions,'' in
  \emph{Proceedings of the IEEE Conference on Computer Vision and Pattern
  Recognition}, 2018, pp. 8601--8610.

\bibitem{moreau2022lens}
A.~Moreau, N.~Piasco, D.~Tsishkou, B.~Stanciulescu, and A.~de~La~Fortelle,
  ``Lens: Localization enhanced by nerf synthesis,'' in \emph{Conference on
  Robot Learning}.\hskip 1em plus 0.5em minus 0.4em\relax PMLR, 2022, pp.
  1347--1356.

\bibitem{chen2017deep}
Z.~Chen \emph{et~al.}, ``Deep learning features at scale for visual place
  recognition,'' in \emph{ICRA}.\hskip 1em plus 0.5em minus 0.4em\relax IEEE,
  2017, pp. 3223--3230.

\bibitem{revaud2019learning}
J.~Revaud, J.~Almaz{\'a}n, R.~S. Rezende, and C.~R.~d. Souza, ``Learning with
  average precision: Training image retrieval with a listwise loss,'' in
  \emph{Proceedings of the IEEE International Conference on Computer Vision},
  2019, pp. 5107--5116.

\bibitem{thoma2020geometrically}
J.~Thoma, D.~P. Paudel, A.~Chhatkuli, and L.~Van~Gool, ``Geometrically mappable
  image features,'' \emph{IEEE Robotics and Automation Letters}, vol.~5, no.~2,
  pp. 2062--2069, 2020.

\bibitem{torii2019large}
A.~Torii, H.~Taira, J.~Sivic, M.~Pollefeys, M.~Okutomi, T.~Pajdla, and
  T.~Sattler, ``Are large-scale 3d models really necessary for accurate visual
  localization?'' \emph{IEEE Transactions on Pattern Analysis and Machine
  Intelligence}, 2019.

\bibitem{li2012worldwide}
Y.~Li, N.~Snavely, D.~Huttenlocher, and P.~Fua, ``Worldwide pose estimation
  using 3d point clouds,'' in \emph{European conference on computer
  vision}.\hskip 1em plus 0.5em minus 0.4em\relax Springer, 2012, pp. 15--29.

\bibitem{liu2017efficient}
L.~Liu, H.~Li, and Y.~Dai, ``Efficient global 2d-3d matching for camera
  localization in a large-scale 3d map,'' in \emph{Proceedings of the IEEE
  International Conference on Computer Vision}, 2017, pp. 2372--2381.

\bibitem{taira2018inloc}
H.~Taira, M.~Okutomi, T.~Sattler, M.~Cimpoi, M.~Pollefeys, J.~Sivic, T.~Pajdla,
  and A.~Torii, ``Inloc: Indoor visual localization with dense matching and
  view synthesis,'' in \emph{Proceedings of the IEEE Conference on Computer
  Vision and Pattern Recognition}, 2018, pp. 7199--7209.

\bibitem{sattler2016efficient}
T.~Sattler, B.~Leibe, and L.~Kobbelt, ``Efficient \& effective prioritized
  matching for large-scale image-based localization,'' \emph{IEEE transactions
  on pattern analysis and machine intelligence}, vol.~39, no.~9, pp.
  1744--1756, 2016.

\bibitem{brachmann2018learning}
E.~Brachmann and C.~Rother, ``Learning less is more-6d camera localization via
  3d surface regression,'' in \emph{Proceedings of the IEEE Conference on
  Computer Vision and Pattern Recognition}, 2018, pp. 4654--4662.

\bibitem{brachmann2017dsac}
E.~Brachmann, A.~Krull, S.~Nowozin, J.~Shotton, F.~Michel, S.~Gumhold, and
  C.~Rother, ``Dsac-differentiable ransac for camera localization,'' in
  \emph{Proceedings of the IEEE conference on computer vision and pattern
  recognition}, 2017, pp. 6684--6692.

\bibitem{kendall2016modelling}
A.~Kendall and R.~Cipolla, ``Modelling uncertainty in deep learning for camera
  relocalization,'' in \emph{2016 IEEE international conference on Robotics and
  Automation (ICRA)}.\hskip 1em plus 0.5em minus 0.4em\relax IEEE, 2016, pp.
  4762--4769.

\bibitem{kendall2017geometric}
------, ``Geometric loss functions for camera pose regression with deep
  learning,'' in \emph{Proceedings of the IEEE conference on computer vision
  and pattern recognition}.\hskip 1em plus 0.5em minus 0.4em\relax IEEE, 2017,
  pp. 5974--5983.

\bibitem{naseer2017deep}
T.~Naseer and W.~Burgard, ``Deep regression for monocular camera-based 6-dof
  global localization in outdoor environments,'' in \emph{2017 IEEE/RSJ
  International Conference on Intelligent Robots and Systems (IROS)}.\hskip 1em
  plus 0.5em minus 0.4em\relax IEEE, 2017, pp. 1525--1530.

\bibitem{valada2018deep}
A.~Valada, N.~Radwan, and W.~Burgard, ``Deep auxiliary learning for visual
  localization and odometry,'' in \emph{2018 IEEE international conference on
  robotics and automation (ICRA)}.\hskip 1em plus 0.5em minus 0.4em\relax IEEE,
  2018, pp. 6939--6946.

\bibitem{clark2017vidloc}
R.~Clark, S.~Wang, A.~Markham, N.~Trigoni, and H.~Wen, ``Vidloc: A deep
  spatio-temporal model for 6-dof video-clip relocalization,'' in
  \emph{Proceedings of the IEEE Conference on Computer Vision and Pattern
  Recognition}, 2017, pp. 6856--6864.

\bibitem{walch2017image}
F.~Walch, C.~Hazirbas, L.~Leal-Taixe, T.~Sattler, S.~Hilsenbeck, and
  D.~Cremers, ``Image-based localization using lstms for structured feature
  correlation,'' in \emph{Proceedings of the IEEE International Conference on
  Computer Vision}, 2017, pp. 627--637.

\bibitem{melekhov2017image}
I.~Melekhov, J.~Ylioinas, J.~Kannala, and E.~Rahtu, ``Image-based localization
  using hourglass networks,'' in \emph{Proceedings of the IEEE international
  conference on computer vision workshops}, 2017, pp. 879--886.

\bibitem{radwan2018vlocnet++}
N.~Radwan, A.~Valada, and W.~Burgard, ``Vlocnet++: Deep multitask learning for
  semantic visual localization and odometry,'' \emph{IEEE Robotics and
  Automation Letters}, vol.~3, no.~4, pp. 4407--4414, 2018.

\bibitem{saha2018improved}
S.~Saha, G.~Varma, and C.~Jawahar, ``Improved visual relocalization by
  discovering anchor points,'' \emph{British Machine Vision Conference}, 2018.

\bibitem{sarlin2021back}
P.-E. Sarlin, A.~Unagar, M.~Larsson, H.~Germain, C.~Toft, V.~Larsson,
  M.~Pollefeys, V.~Lepetit, L.~Hammarstrand, F.~Kahl \emph{et~al.}, ``Back to
  the feature: Learning robust camera localization from pixels to pose,'' in
  \emph{Proceedings of the IEEE/CVF Conference on Computer Vision and Pattern
  Recognition}, 2021, pp. 3247--3257.

\bibitem{yang2019sanet}
L.~Yang, Z.~Bai, C.~Tang, H.~Li, Y.~Furukawa, and P.~Tan, ``Sanet: Scene
  agnostic network for camera localization,'' in \emph{Proceedings of the
  IEEE/CVF International Conference on Computer Vision}, 2019, pp. 42--51.

\bibitem{gordo2017end}
A.~Gordo, J.~Almazan, J.~Revaud, and D.~Larlus, ``End-to-end learning of deep
  visual representations for image retrieval,'' \emph{International Journal of
  Computer Vision}, vol. 124, no.~2, pp. 237--254, 2017.

\bibitem{radenovic2018fine}
F.~Radenovi{\'c}, G.~Tolias, and O.~Chum, ``Fine-tuning cnn image retrieval
  with no human annotation,'' \emph{IEEE transactions on pattern analysis and
  machine intelligence}, vol.~41, no.~7, pp. 1655--1668, 2018.

\bibitem{thoma2020soft}
J.~Thoma, D.~P. Paudel, and L.~Van~Gool, ``Soft contrastive learning for visual
  localization,'' \emph{Advances in Neural Information Processing Systems 33},
  2020.

\bibitem{chen2014convolutional}
Z.~Chen, O.~Lam, A.~Jacobson, and M.~Milford, ``Convolutional neural
  network-based place recognition,'' \emph{preprint arXiv:1411.1509}, 2014.

\bibitem{sermanet2014overfeat}
P.~Sermanet, D.~Eigen, X.~Zhang, M.~Mathieu, R.~Fergus, and Y.~LeCun,
  ``Overfeat: Integrated recognition, localization and detection using
  convolutional networks,'' in \emph{2nd International Conference on Learning
  Representations, ICLR 2014}, 2014.

\bibitem{sunderhauf2015performance}
N.~S{\"u}nderhauf, S.~Shirazi, F.~Dayoub, B.~Upcroft, and M.~Milford, ``On the
  performance of convnet features for place recognition,'' in
  \emph{IROS}.\hskip 1em plus 0.5em minus 0.4em\relax IEEE, 2015, pp.
  4297--4304.

\bibitem{leyva2021generalized}
M.~Leyva-Vallina, N.~Strisciuglio, and N.~Petkov, ``Generalized contrastive
  optimization of siamese networks for place recognition,'' \emph{arXiv
  preprint arXiv:2103.06638}, 2021.

\bibitem{hausler2019multi}
S.~Hausler, A.~Jacobson, and M.~Milford, ``Multi-process fusion: Visual place
  recognition using multiple image processing methods,'' \emph{IEEE Robotics
  and Automation Letters}, vol.~4, no.~2, pp. 1924--1931, 2019.

\bibitem{hausler2020hierarchical}
S.~Hausler and M.~Milford, ``Hierarchical multi-process fusion for visual place
  recognition,'' in \emph{2020 IEEE International Conference on Robotics and
  Automation (ICRA)}.\hskip 1em plus 0.5em minus 0.4em\relax IEEE, 2020, pp.
  3327--3333.

\bibitem{mescheder2019occupancy}
L.~Mescheder, M.~Oechsle, M.~Niemeyer, S.~Nowozin, and A.~Geiger, ``Occupancy
  networks: Learning 3d reconstruction in function space,'' in
  \emph{Proceedings of the IEEE/CVF Conference on Computer Vision and Pattern
  Recognition}, 2019, pp. 4460--4470.

\bibitem{oechsle2019texture}
M.~Oechsle, L.~Mescheder, M.~Niemeyer, T.~Strauss, and A.~Geiger, ``Texture
  fields: Learning texture representations in function space,'' in
  \emph{Proceedings of the IEEE/CVF International Conference on Computer
  Vision}, 2019, pp. 4531--4540.

\bibitem{niemeyer2020differentiable}
M.~Niemeyer, L.~Mescheder, M.~Oechsle, and A.~Geiger, ``Differentiable
  volumetric rendering: Learning implicit 3d representations without 3d
  supervision,'' in \emph{Proceedings of the IEEE/CVF Conference on Computer
  Vision and Pattern Recognition}, 2020, pp. 3504--3515.

\bibitem{mildenhall2020nerf}
B.~Mildenhall, P.~P. Srinivasan, M.~Tancik, J.~T. Barron, R.~Ramamoorthi, and
  R.~Ng, ``Nerf: Representing scenes as neural radiance fields for view
  synthesis,'' in \emph{European conference on computer vision}.\hskip 1em plus
  0.5em minus 0.4em\relax Springer, 2020, pp. 405--421.

\bibitem{torii201524}
A.~Torii, R.~Arandjelovic, J.~Sivic, M.~Okutomi, and T.~Pajdla, ``24/7 place
  recognition by view synthesis,'' in \emph{Proceedings of the IEEE Conference
  on Computer Vision and Pattern Recognition}, 2015, pp. 1808--1817.

\bibitem{zhang2021reference}
Z.~Zhang, T.~Sattler, and D.~Scaramuzza, ``Reference pose generation for
  long-term visual localization via learned features and view synthesis,''
  \emph{International Journal of Computer Vision}, vol. 129, no.~4, pp.
  821--844, 2021.

\bibitem{yen2021inerf}
L.~Yen-Chen, P.~Florence, J.~T. Barron, A.~Rodriguez, P.~Isola, and T.-Y. Lin,
  ``inerf: Inverting neural radiance fields for pose estimation,'' in
  \emph{2021 IEEE/RSJ International Conference on Intelligent Robots and
  Systems (IROS)}.\hskip 1em plus 0.5em minus 0.4em\relax IEEE, 2021, pp.
  1323--1330.

\bibitem{glocker2013real}
B.~Glocker, S.~Izadi, J.~Shotton, and A.~Criminisi, ``Real-time rgb-d camera
  relocalization,'' in \emph{2013 IEEE International Symposium on Mixed and
  Augmented Reality (ISMAR)}.\hskip 1em plus 0.5em minus 0.4em\relax IEEE,
  2013, pp. 173--179.

\bibitem{hendrycks2016gaussian}
D.~Hendrycks and K.~Gimpel, ``Gaussian error linear units (gelus),''
  \emph{arXiv preprint arXiv:1606.08415}, 2016.

\bibitem{brachmann2019expert}
E.~Brachmann and C.~Rother, ``Expert sample consensus applied to camera
  re-localization,'' in \emph{Proceedings of the IEEE/CVF International
  Conference on Computer Vision}, 2019, pp. 7525--7534.

\bibitem{sarlin2019coarse}
P.-E. Sarlin, C.~Cadena, R.~Siegwart, and M.~Dymczyk, ``From coarse to fine:
  Robust hierarchical localization at large scale,'' in \emph{CVPR}, 2019, pp.
  12\,716--12\,725.

\bibitem{newcombe2011kinectfusion}
R.~A. Newcombe, S.~Izadi, O.~Hilliges, D.~Molyneaux, D.~Kim, A.~J. Davison,
  P.~Kohi, J.~Shotton, S.~Hodges, and A.~Fitzgibbon, ``Kinectfusion: Real-time
  dense surface mapping and tracking,'' in \emph{2011 10th IEEE international
  symposium on mixed and augmented reality}.\hskip 1em plus 0.5em minus
  0.4em\relax IEEE, 2011, pp. 127--136.

\bibitem{he2016deep}
K.~He, X.~Zhang, S.~Ren, and J.~Sun, ``Deep residual learning for image
  recognition,'' in \emph{Proceedings of the IEEE conference on computer vision
  and pattern recognition}, 2016, pp. 770--778.

\bibitem{berton2021viewpoint}
G.~Berton, C.~Masone, V.~Paolicelli, and B.~Caputo, ``Viewpoint invariant dense
  matching for visual geolocalization,'' in \emph{Proceedings of the IEEE/CVF
  International Conference on Computer Vision}, 2021, pp. 12\,169--12\,178.

\end{thebibliography}
}

\end{document}